\documentclass{article}

\usepackage{arxiv}

\usepackage[utf8]{inputenc} %
\usepackage[T1]{fontenc}    %
\usepackage{hyperref}       %
\usepackage{url}            %
\usepackage{booktabs}       %
\usepackage{amsfonts}       %
\usepackage{nicefrac}       %
\usepackage{microtype}      %
\usepackage{cleveref}       %
\usepackage{lipsum}         %
\usepackage{graphicx}
\usepackage{natbib}
\usepackage{doi}

\usepackage[whole]{bxcjkjatype}
\usepackage{graphicx}
\usepackage{cprotect}
\usepackage{booktabs}
\usepackage{longtable}
\usepackage{chngcntr}
\usepackage{color}    %
\usepackage{subfig}

\title{COMPASS: a Creative Support System that \\Alerts Novelists to the Unnoticed Missing Contents}

\author{Yusuke Mori$^{1}$ \, Hiroaki Yamane$^{2, 1}$ \, Ryohei Shimizu$^{1}$ \, Yusuke Mukuta$^{1,2}$ \, Tatsuya Harada$^{1,2}$  \\
         $^{1}$The University of Tokyo,  $^{2}$RIKEN \\
         {\tt \{mori, yamane, shimizu, mukuta, harada\}@mi.t.u-tokyo.ac.jp}
         }

\renewcommand{\shorttitle}{COMPASS: a Creative Support System}

\hypersetup{
pdftitle={COMPASS: a Creative Support System that Alerts Novelists to the Unnoticed Missing Contents},
pdfsubject={},
pdfauthor={Yusuke Mori, Hiroaki Yamane, Ryohei Shimizu, Yusuke Mukuta, Tatsuya Harada},
pdfkeywords={creative-support-system},
}

\begin{document}

\maketitle

\begin{abstract}
When humans write, they may unintentionally omit some information. Complementing the omitted information using a computer is helpful in providing writing support. Recently, in the field of story understanding and generation, story completion (SC) was proposed to generate the missing parts of an incomplete story. Although its applicability is limited because it requires that the user have prior knowledge of the missing part of a story, missing position prediction (MPP) can be used to compensate for this problem. MPP aims to predict the position of the missing part, but the prerequisite knowledge that ``one sentence is missing'' is still required. In this study, we propose Variable Number MPP (VN-MPP), a new MPP task that removes this restriction; that is, the task to predict multiple missing sentences or to judge whether there are no missing sentences in the first place. We also propose two methods for this new MPP task. Furthermore, based on the novel task and methods, we developed a creative writing support system, \textbf{COMPASS}. The results of a user experiment involving professional creators who write texts in Japanese confirm the efficacy and utility of the developed system. 
\end{abstract}


\section{Introduction}
\label{sec:intro}

\subsection{Story Understanding and Story Generation}
\label{section:introduction}

Creativity is human nature. 
Writing and reading stories are essential aspects of creativity.
Moreover, understanding how humans write and interpret stories is inextricably linked to understanding humans themselves.
\citet{Winston_2011_AAAI_Hypotheses} theorized this as \textit{The Strong Story Hypothesis}: The mechanisms that enable humans to tell, understand, and recombine stories separate human intelligence from that of other primates. 
Stories are used not only for entertainment, but also for a variety of other purposes: advertisement \citep{ono-etal-2019-advertising}, marketing \citep{McKee_2018_Storynomics}, education \citep{Theme-Rewriting}, and serious storytelling \citep{Lugmayr_2017_serious_storytelling}.
Storytelling is deeply rooted in human life.

Currently, thanks to the Internet, anybody can freely publish their original stories.
However, creating a story is not an easy task. It is even more difficult to write something that people like and want to read. 
Sometimes, even professional writers fall into slumps (so-called ``writer's block'') during the writing process and cannot complete their stories.
To understand the secret of creating good stories, various studies have been conducted \citep{Campbell1949}. 
Here, to clarify what it means to create a story, it is important to have the practical knowledge of creators who are actually engaged in the creation process.
In fact, it is true that the knowledge of many creators helps in understanding a story \citep{Forster1927,Propp1928,Vonnegut1981_video}. 
Rules for creating stories have been extensively studied; ``Three-act structure''~\citep{SydField1984} and ``Save the cat''~\citep{savethecat} are famous examples. 
These works can help guide people who wish to create good stories to demonstrate their creativity.
In this way, story understanding and story generation are inextricably linked.

From the viewpoint of aiming for computational creativity, automatic storytelling or narrative generation has been an important step \citep{Gervas_2009_storytelling_and_creativity,garbacea2020neural}.
The notion of automatic storytelling has always fascinated researchers, and models appeared as early as the 1970s \citep{Klein1973AutomaticNW,meehan1976metanovel,Meehan_1977_IJCAI,LEBOWITZ1984171,LEBOWITZ1985483}. 

With the development of machine learning (ML) and natural language processing (NLP) technology in recent years, automatic storytelling models have made great strides \citep{WritingPromptsDataset,fan-etal-2019-strategies,Guan_2020_TACL,MEGATRON-CNTRL}.
The creation of an automated system that can support the creative endeavors of people is now feasible \citep{roemmele_writing_2016,Clark_2018_creative_Writing_machine_in_the_loop,peng2018towards,yao2019plan,goldfarb2019plan}. 

\subsection{Story Completion for Creative Support}
\label{section:sc_for_creative_support}

It is essential to train computers to understand and create stories, and thereby assist people in the story-creation process.
To measure the reading comprehension abilities of systems regarding stories, \citet{mostafazadeh2016} proposed the ``Story Cloze Test'' (SCT). In the SCT, four sentences are presented, and the last sentence is excluded from an original five-sentence story. The objective is to select an appropriate last sentence from two options.
This research on story understanding spread to the research on story generation, partly due to the usefulness of the dataset ``ROCStories'' proposed at the same time.
As a typical example, ``Story Ending Generation'' (SEG) was designed as a subtask of story generation \citep{Zhao2018plotstoendings}. 
Further, based on SEG, Wang and Wan \citep{wang_tcvae} proposed a ``Story Completion'' (SC) task. Given any four sentences of a five-sentence story, the task's objective is to generate the sentence that is not given (referred to as the missing plot) to complete the story.
The ability to solve the story completion task is important in the context of creative support. 
If writers cannot complete the story and do not know how to continue the plot, a suitable model could provide appropriate support.

Here, to clarify the position of our research, we divide ``story generation'' into two main groups---\textbf{Open-ended story generation} and \textbf{Context-aware story generation}.\footnote{Note that this classification is not absolute and other classifications can be considered. For example, \citet{Hou_2019_survey_dl_storygeneration} classified story generation models into three types according to the types of conditions (user constraints) given at the time of generation: Theme-Oriented models, Storyline-Oriented models, and Human--Machine Interaction-Oriented models}

Open story generation is a problem proposed by \citet{Li_2013_AAAI}, in which a story about any domain is automatically generated without a priori manual knowledge engineering.
The intent of the authors of each paper may differ, but it seems the terms ``open-domain story generation'' and ``open-ended story generation'' \citep{UNION} can be included in this group.\footnote{We were hesitant about including ``open-domain'' here, but Wang and Wan \citep{wang_tcvae}, who proposed the story completion task, wrote ``Our future work will focus on story completion and story generation task in open-domain.'' Hence, we think that open-domain story generation can be separated from the story completion task.} 
Because it seems that ``open-ended'' is easily understood and shows the general meaning of ``not restricted,'' for the sake of simplicity, we will refer to story generation in this direction as ``open-ended story generation'' in this paper.
Note that giving a prompt/title as an input to the models is generally used in open-ended storytelling. This kind of input is not considered a ``restriction'' in this area.

We define ``Context-aware story generation'' as story generation tasks in which a human-written story/plot (so-called ``context'') is given as an input and, according to the given context, models generate subsequent sentences, complementary sentences in the missing middle part, etc.
Typical examples of this approach include SEG \citep{Zhao2018plotstoendings}, SC \citep{wang_tcvae}, and story infilling \citep{ippolito-etal-2019-unsupervised}.

In open-ended story generation, machine-generated stories are variables, but this does not mean that open-ended story generation is more difficult than context-aware story generation. For example, in the story completion task, it is necessary for models to understand the context that has already been written by humans and then fill in the gaps. It needs both aspects of story understanding and story generation, thus restriction by context does not mean the task is easy.
Both open-ended story generation and context-aware story generation have important implications. 
With that in mind, in this study, we focus on the latter approach.

To overcome the issue of conventional story completion tasks requiring information regarding the position of the missing part in a story, we previously proposed ``Missing Position Prediction'' (MPP) as a task to predict the position based on the given incomplete story. In our previous paper \citep{mori-etal-2020-finding}, we proposed MPP with limited conditions (LC-MPP). In this paper, referring to LC-MPP, we propose an updated version of MPP, Various Number MPP (VN-MPP), as a task closer to a more realistic setting.

In story and narrative research, it is necessary to define a story and the kind of text that can be regarded as a story; the task of judging whether a text is a story is known as story detection \citep{eisenberg-finlayson:2017:EMNLP2017}. 
We define a story as a series of events related to characters and having a beginning and an end; these events are intended to change the emotions and relationships of the characters.

\subsection{Summary of Contributions}
\label{sec:summary_of_contributions}

The major contributions of this study are summarized as follows.

\begin{itemize}
    \item To overcome the issue of conventional SC tasks requiring information regarding the position of the missing part in a story, we previously proposed ``Missing Position Prediction'' as a task to predict the position based on the given incomplete story. At first, we proposed MPP with limited conditions (LC-MPP) \citep{mori-etal-2020-finding}. However, in this paper, referring to LC-MPP, we propose an updated version of MPP, Various Number MPP (VN-MPP), as a task closer to a more realistic setting.
    \item We propose two novel methods for VN-MPP and Story Completion (SC): the two-module approach and the end-to-end approach.
    \item Based on our proposed tasks and methods, we developed a system for human story writing assistance. We named this system \textbf{``COMPASS''}, which stands for a writing support system to \textbf{COMP}lement \textbf{A}uthor unaware \textbf{S}tory gap\textbf{S}. Subsequently, four professionals in the field of creative writing in Japanese evaluated the developed system and confirmed its efficacy and utility.
\end{itemize}

This paper is divided into two major parts.
In the first part, we present the proposed VN-MPP task (Section \ref{sec:task_description}) and a proposed method to solve it (Section \ref{sec:methodology}).
In the second part, we discuss the implementation of VN-MPP as a creation support system (Section \ref{sec:demonstration-system}) and the experimental verification of its practicality by professionals (Section \ref{sec:vnmpp-user-study}).

We plan to make the source codes publicly available in the future.

\section{Related Work}
\label{sec:related_work}

\subsection{Writing Assistance by Text Generation}
\label{subsec:story_writing_assistance}

Applying machine learning to human story writing assistance is an approach of which interesting works were published in recent years \citep{roemmele_writing_2016,peng2018towards,yao2019plan,goldfarb2019plan}.
Referring to Recurrent Neural Networks (RNN) as a promising machine learning framework for language generation tasks, \citet{roemmele_writing_2016} envisioned the task of narrative auto-completion applied to helping an author write a story. 
\citet{peng2018towards} proposed an analyze-to-generate framework for controllable story generation. They apply two types of generation control: 1) ending valence control (happy or sad ending) and 2) storyline keywords.
\citet{yao2019plan} proposed a two-step pipeline for open-domain story generation: 1) story planning, which generates a storyline represented by an ordered list of words, and 2) surface realization, which composes a story based on the storyline. They proposed a hierarchical generation framework named \textit{plan-and-write} that combines storyline planning and surface realization to generate stories from titles.
Based on the studies carried out by \citet{yao2019plan} and \citet{holtzman-etal-2018-learning}, \citet{goldfarb2019plan} presented a neural narrative generation system named \textit{Plan-and-Revise} in which humans and computers collaborate to generate stories.

This research is positioned in the context of such research on creative writing support, and at the same time aims to apply the research on SC to creative writing support.

\subsection{Story Understanding and Generation}
\label{subsection:story_understanding_and_generation}

SC was recently proposed in the field of story understanding and generation as a method for generating the missing parts of an incomplete story.
Here, we discuss the research on story understanding and generation, which are strongly associated with SC.

At the intersection of NLP and literary analysis, various studies have been conducted.
Referring to the narrative cloze test \citep{NarrativeClozeTest} as a typical example of a story understanding task considering events, \citet{mostafazadeh2016} proposed SCT as a more difficult task. SCT presents four sentences, and the last sentence is excluded from a story composed of five sentences. The system must select an appropriate sentence from two choices that complement the missing last sentence.
In addition to the task, the authors released a large-scale story corpus named ROCStories, \footnote{\url{https://www.cs.rochester.edu/nlp/rocstories/}}
which is a collection of non-fictional daily-life stories written by hundreds of workers at Amazon Mechanical Turk. The five-sentence stories contain varied common-sense knowledge.
SCT was proposed as a challenging task, but the improvement of NLP methods is so rapid that an even more challenging task was needed to evaluate the performance of models in this ever-evolving field.
When SCT was proposed, machine learning models could solve this task with an accuracy of less than 60\%. 
However, in a few years, \citet{chaturvedi-peng-roth:2017:EMNLP2017} achieved an accuracy of 77.6\% and \citet{Radford2018} achieved 86.5\%.

With the development of the field of story understanding, the field of story generation has also become more active in research.
Story generation approaches can be roughly divided into two types. One involves studies that produce the entire story \citep{WritingPromptsDataset}. The other involves studies that complement the existing incomplete text \citep{ippolito-etal-2019-unsupervised,donahue-etal-2020-enabling,huang-etal-2020-inset,wang2020narrative}. 
In this paper, we focus on the latter approach. This is because we believe that generating sentences to improve an incomplete story is indeed an important task in human writing assistance.

\subsection{Story Completion and Missing Position Prediction}
\label{subsection:related_work_SC_and_MPP}

Inspired by the aforementioned task, SCT, which is a subtask of story generation, SEG was designed by \citet{Zhao2018plotstoendings}. 
In their SEG, a system is given an incomplete story, where the last sentence is excluded from the original five-sentence story. The objective of the task is to automatically generate the last sentence of this given incomplete story.
Furthermore, based on SEG, \citet{wang_tcvae} proposed an SC task and investigated the problem of generating missing story plots at any position in an incomplete story. 
If a sentence in the middle is missing, the task becomes more difficult because the system must capture the context both before and after the missing sentence.
Additionally, in recent years, research regarding text infilling has been actively conducted \citep{ippolito-etal-2019-unsupervised,donahue-etal-2020-enabling,huang-etal-2020-inset}. Regarding stories, \citet{ippolito-etal-2019-unsupervised} worked on complementing the missing span between left and right contexts, which they called ``story infilling.'' 

However, these studies require that the writer have prior knowledge of the missing parts, and they do not consider the case where the writer is unaware of the flaws in their work.
To overcome this limitation, we \citep{mori-etal-2020-finding} proposed a story comprehension task named MPP.
In MPP, an incomplete story with one sentence missing is given as input. Unlike the previously mentioned task, no information regarding the position of the missing content is required. MPP requires the prediction of the position of the missing part.
The ability to solve this task indicates that computers can identify flaws in a story's plot.
Referring to our previous work \citep{mori-etal-2020-finding}, \citet{cai2020narrative} pointed out that most previous studies focused solely on the problem of what to infill/modify and the need to know the positions of missing parts a priori. 
To solve MPP, it is necessary to identify unnaturalness because 
MPP is deeply related to a fundamental question in story understanding: whether or not the model understands the flow of a story.

However, conventional MPP has several restrictions.
In the first MPP approach proposed by us \citep{mori-etal-2020-finding}, it must be known a priori that there is a missing position in the input story and that there is only one such instance.
In reality, an input story may be complete, i.e., it has no missing parts. Furthermore, there may be a case in which there are multiple missing positions.
In this paper, we propose VN-MPP in which the number of missing positions, including zero, can vary. 

\subsection{Neural Network Language Models}
\label{related_work:language_model}

Referring to RNNs as a promising machine learning framework for language generation tasks, \citet{roemmele_writing_2016} envisioned the task of narrative auto-completion applied to helping an author write a story.
With the advent of the sequence-to-sequence model (Seq2seq), the use of neural networks as a method for generating natural sentences has become commonplace.
Seq2seq was first proposed for machine translation \citep{Sutskever:2014:SSL:2969033.2969173}. However, it has been widely applied to other tasks in NLP \citep{Vinyals2015conversation}.
In SEG, simple Seq2seq and an extension using the attention mechanism are used as a baseline \citep{Zhao2018plotstoendings,li-etal-2018-generating,Guan2019,mori-etal-2019-toward}.

Transformer \citep{Vaswani_2017_transformer}, which replaced the RNNs in Seq2seq with self-attention, is the basis of today's significant improvement in NLP.
Unsupervised pre-trained large neural models, such as BERT \citep{devlin-etal-2019-bert} and GPT-2 \citep{radford2019language}, were proposed using the Transformer architecture and soon became the mainstream in NLP.
These pre-trained models were roughly divided into two: one uses the transformer encoder (bi-directional architecture) and the other uses the transformer decoder (left-to-right architecture).
In sequence generation, it was common knowledge that models using left-to-right architecture \citep{radford2019language,Yang_NIPS2019_8812} are more suitable. However, instead of using only one of the transformer encoder based architecture or the transformer decoder based architecture, attempts to create Seq2seq (Encoder-Decoder) models using unsupervised pre-trained large neural models for initializing each of the encoders and the decoders are becoming the new mainstream \citep{lewis-etal-2020-bart,Rothe2020}.
In this paper, our proposed method is based on BART (proposed by \citet{lewis-etal-2020-bart}), which uses BERT as the encoder and GPT-2 as the decoder, and it exhibits high performance in tasks, such as summarization.

\subsection{Evaluation Metrics}
\label{subsection:related_evaluation_metrics}

In text generation tasks, human evaluation, i.e., to involve humans as judges, is generally thought of as the gold standard. This is not unnatural, because most of the text generation models aim to generate ``natural'' text for humans.
However, some problems remain.
Human evaluation is costly, time-consuming, and dependent on individual abilities. 
Regarding Amazon Mechanical Turk, which is a generally used crowdsourcing platform, \citet{ippolito-etal-2019-unsupervised} reported that the evaluation by average workers is unreliable in the task of story infilling. They inserted one honeypot question in 11 questions and found that performance on the honeypot question was close to random guessing. 
\citet{august-etal-2020-exploring} pointed out that human evaluation schemes tend to ignore the difference of perspectives, authors, and readers. 

Therefore, automatic metrics to evaluate the day-by-day progress of natural language generation are strongly needed.
However, it has been shown that traditional metrics have poor correlation with human evaluation, so proper evaluation of text generation is difficult \citep{liu-EtAl:2016:EMNLP20163,novikova-etal-2017-need,chaganty-etal-2018-price,Gatt_and_Krahmer_suvery_NLG,hashimoto-etal-2019-unifying}. 
Recently, based on large unsupervised pre-trained neural models, various machine-learned metrics have been proposed and tested: BERTScore \citep{bert-score} and BLEURT \citep{sellam-etal-2020-bleurt}.
In particular, for story generation, \citet{UNION} proposed UNION, a learnable \textit{UNreferenced metrIc for evaluating Open-eNded story generation}. 

Based on the above metrics, we evaluated how well the method proposed in this paper performed on our proposed VN-MPP task.

\section{Task Description}
\label{sec:task_description}
We begin by formulating SEG, SC, and MPP. Then, we formulate our proposed VN-MPP.

\subsection{Story Ending Generation and Story Completion}
\label{subsec:task_seg_sc}

We define $ S = \{s_1, s_2, ..., s_n\} $ as a story comprising $n$ sentences.
In SEG, $S' = \{s_1, s_2, ..., s_{n-1}\} $ is given as an input. 
The objective of the task is to generate an appropriate ending.
For story completion, an incomplete story consisting of $n - 1$ sentences $S' = \{s_1, ..., s_{k-1}, s_{k+1}, ..., s_{n}\} $, where $k$ represents the position of the missing sentence in the story, is provided.
Next, the objective of the task is to generate an appropriate sentence that is coherent with the given sentences.
During each task, the model is trained to maximize probability $p(y|S')$, where $y$ represents the ground truth sentence.

\subsection{Conventional Missing Position Prediction (MPP)}
\label{subsec:task_MPP}

To overcome the issue of the story completion model requiring information about $k$, i.e., the position of the missing sentence, our previous work \citep{mori-etal-2020-finding} proposed MPP to predict $k$ from a given $n - 1$ sentences.
Similar to the story completion task, an incomplete story comprising $n - 1$ sentences $S' = \{s_1, ..., s_{k-1}, s_{k+1}, ..., s_{n}\} $ is given as an input. However, no information about $k$ is provided.
The order of the sentences is known, but the missing position is unknown.
More specifically, $s_{k-1}$ and $s_{k+1}$ are treated as continuous sentences.
Here, their objective is to predict $k$ from the input. In other words, the model is trained to maximize probability $p(missing=k|S')$. 

\subsection{Variable Number Missing Position Prediction (VN-MPP)}
\label{subsec:task_vn-mpp}

In MPP, it should be known a priori that there is a missing position in the input story and that there is only one such instance.
In this study, we propose VN-MPP in which the number of missing positions, including zero, can be variable.

A story comprising $n-m$ ($0 \leq m < n$) sentences $S' = \{s_{i_1}, s_{i_2}, ..., s_{i_j}, s_{i_{j+1}}, ..., s_{i_{n-m}}\}$ is given as an input. However, no information about $m$ is provided. We use $i, j$ to indicate that for a given incomplete story, we do not know how many sentences are missing from the original text. For example, we may be given the first, third, and fifth sentences (i.e. $\{s_1, s_3, s_5\}$) as $S'$, but do not know their original position, hence they are represented as the first, second, and third sentences in the incomplete story (i.e. $\{s_{i_1}, s_{i_2}, s_{i_3}\}$).
The order of the sentences is known, but the number of missing positions and the location of each missing position are unknown.
More specifically, $s_{i_j}$ and $s_{i_{j+1}}$ are treated as continuous sentences, but there may exist some lost sentences.
Here, our objective is to predict all the missing positions from the input, including the case where there is no missing position ($m=0$), i.e., where the input story is complete.  

We note here that even when there is a missing part in the story, it may be caused by a writer's intention in that ``I want the readers to read between lines.'' However, the missing part can also be an unintentional mistake. 
To analyze if the model can understand whether the missing part is a ``writer's intentional omission'' is out of the scope of this study. 

\section{Methodology: two basic approaches}
\label{sec:methodology}

To solve our proposed VN-MPP and SC, we propose two methods in this section. 
Then, in Section \ref{sec:demonstration-system}, we arrange one of these methods to be more suitable for a creative writing assistance system.

\begin{itemize}
    \item Two approaches for VN-MPP
    \begin{itemize}
        \item \textbf{Two-module Approach}
        \item \textbf{End-to-end Approach}
    \end{itemize}
    \item Arranged method (will be discussed in Section \ref{sec:demonstration-system})
    \begin{itemize}
        \item \textbf{Two-module Approach with improved SC module (Two-module v2)}
    \end{itemize}
\end{itemize}

As mentioned earlier, we will first discuss the following two approaches: \textbf{Two-module Approach} and \textbf{End-to-end Approach}.

The first method consists of two modules: the VN-MPP module and the SC module. In this study, we call this approach the two-module approach.
The second method treats VN-MPP and SC in an end-to-end manner. In this study, we call this approach the end-to-end approach.
We also provide the details of dataset preprocessing, which is another key factor of our proposed methodology.

\subsection{Two-module Approach}
\label{sec:two-module-approach}

Although the end-to-end method is considered to be the mainstream in the field of machine learning, in the first method, we deliberately divided the module for each task so that the output of VN-MPP alone could be confirmed.
In addition, because various methods for SC have been proposed and are expected to be developed in the future, it is also advantageous that our proposed VN-MPP module can be easily combined with these methods.

As written above, we make two-modules: the VN-MPP module and the SC module. However, we propose a way to handle these modules through a unified structure method.

We use large transformer-based Seq2seq models to solve both VN-MPP and SC.
In other words, we treat both VN-MPP and SC as Seq2seq type tasks, represented by translation and summarization, etc.
Because the tasks are conversions within the same language, we can consider them as tasks similar to summarization in form.

We introduce a new special token \verb|<missing_sentence>| to solve both VN-MPP and SC.
In VN-MPP, an input is an incomplete story, and the output is an incomplete story with its missing part filled with \verb|<missing_sentence>| tokens.
In SC, an input is an incomplete story with \verb|<missing_sentence>| tokens, and the output is a completed story where \verb|<missing_sentence>| tokens are replaced with appropriate sentences.

\subsection{End-to-end Approach}
\label{sec:end-to-end-approach}

Our end-to-end approach is simpler than our two-module approach.

VN-MPP and SC were handled consistently in this approach and treated as a Seq2seq task.
Because this end-to-end approach also involves conversions within the same language, we can consider it as a task similar to summarization in form.

In this case, there is no need to introduce additional special tokens.
We simply input an incomplete story to the Seq2seq encoder, after which we obtain a complete story as an output from the Seq2seq decoder.

\subsection{Dataset Preprocessing}
\label{sec:dataset_preprocessing}

\begin{figure}[!t]
  \center
  \includegraphics[width=\textwidth]{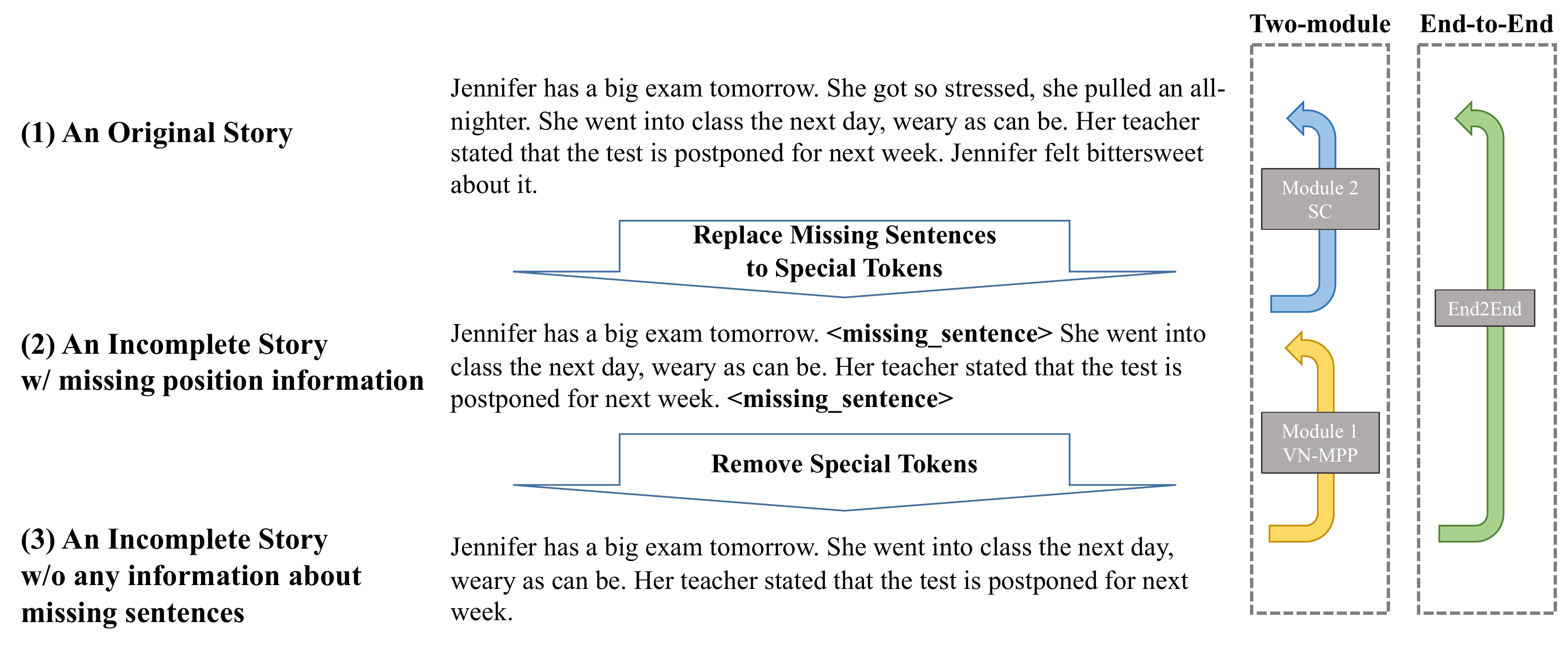}
  \caption{\label{fig:dataset_preprocessing} Conceptual diagram of dataset preprocessing and its correspondence to the proposed methods.}
\end{figure}

From the original complete stories from a dataset, we preprocess them and create two sets of misinformation data. A conceptual diagram of the preprocessing is presented in Figure \ref{fig:dataset_preprocessing}. \footnote{In the actual implementation, we first delete randomly selected ``missing sentences'' and then insert tokens.}

Given an original story comprising $n$ sentences, it can be written as equation \ref{eq:original_story}. 

\begin{equation}
\label{eq:original_story}
	S = \{s_1, s_2, ..., s_n\}
\end{equation}

From this, we create two incomplete stories.
The first one is an incomplete story with \verb|<missing_sentence>| tokens.
We randomly choose the number of missing sentences $m$, after which we replace $m$ sentences from the original story with the special tokens.

\begin{eqnarray}
\label{eq:story_with_special_missing_token}
  S'' &=& \{s_{i_1}, \verb|<missing_sentence>|, s_{i_2}, ..., s_{i_j}, \verb|<missing_sentence>|, \nonumber \\ 
  & & s_{i_{j+1}}, ..., s_{i_{n-m}}\}
\end{eqnarray}

Next, we remove the special tokens and create an incomplete story without any information regarding the missing positions.

\begin{equation}
\label{eq:incomplete_story}
  S'  = \{s_{i_1}, s_{i_2}, ..., s_{i_j}, s_{i_{j+1}}, ..., s_{i_{n-m}}\}   
\end{equation}

The aim of the entire task can be explained as the conversion from (\ref{eq:incomplete_story}) to (\ref{eq:original_story}).
Our end-to-end approach tries to do this in one step.
On the other hand, our two-module approach does two conversions: VN-MPP module convert \ref{eq:incomplete_story} to (\ref{eq:story_with_special_missing_token}), and SC module convert (\ref{eq:story_with_special_missing_token}) to (\ref{eq:original_story}).

How we define $m$ for each dataset is described in detail in Section \ref{sec:dataset}.

\subsection{Implementation Details}
\label{subsec:implementation-details_methodology}

We implement our codes based on PyTorch \citep{PyTorch}, an open-source machine learning framework provided as a Python library.\footnote{\url{https://pytorch.org/}}
To make use of unsupervised pre-trained large neural models, our code is also based on Hugging Face Transformers \citep{wolf-etal-2020-transformers}, which provides general-purpose architectures for Natural Language Understanding (NLU) and Natural Language Generation (NLG) for both TensorFlow 2.0 and PyTorch.
To add details, we use PyTorch version 1.7.0 and Hugging Face Transformers version 4.5.1.

\subsection{Datasets}
\label{sec:dataset}

We use ROCStories as a story dataset and, from other domains, we use the CNN / Daily Mail summarization dataset \citep{see-etal-2017-get} to investigate how useful our proposed VN-MPP is in domains other than stories.

\subsubsection{Story Dataset}

\begin{table}[!th]
\small
\centering
\begin{tabular}{lrr}  
\toprule
set  & \#stories & missing position \\
\midrule
train       & 78,528 & Given randomly during training \\
dev    & 9,816 & Given when creating dataset \\
test   & 9,817 & Given when creating dataset \\
\midrule
total  & 98,161 & \\
\bottomrule
\end{tabular}
\caption{Overview of the ROCStories with our preprocessing.}
\label{tab:dataset_rocstories}
\end{table}

ROCStories is typically used in SEG \citep{Guan2019,li-etal-2018-generating,Zhao2018plotstoendings}. 
Similarly, \citep{wang_tcvae} used ROCStories for their story completion task. 
The dataset is also used for controllable story generation \citep{peng2018towards}.
Even in Counterfactual Story Rewriting, a more advanced story revising task, the \textsc{TimeTravel} dataset proposed for the task is built on top of ROCStories \citep{qin-etal-2019-counterfactual}.
 
It has been pointed out that the Story Cloze Test dataset (SCT-v1.0), published together with ROCStories, has a bias. There is too much difference between a right ending and a wrong ending, so the classification performance can be improved in unintended ways. \citep{sharma-etal-2018-tackling} proposed SCT-v1.5 datasets to avoid the bias. 
However, we use ROCStories, the dataset used for training in SCT, which is not directly related to the SCT-v1.0 bias. In addition, SCT-v1.5 is intended to put restrictions on workers when writing right/wrong endings in order to make the two-choice question adequately difficult. Of course, SCT-v1.5 is superior to SCT-v1.0 in conducting SCT, but we believe that SCT-v1.0 and ROCStories are closer to stories that humans can write freely (although there are some rules including five-sentence restriction).

As shown in Table \ref{tab:dataset_rocstories}, the dataset was randomly split in the ratio of 8:1:1 to obtain the training, development, and test sets, respectively.
We removed sentences from a five-sentence story. The number of missing positions $m$ was randomly decided to be $ 0 \leq m \leq 2 $, based on a discrete uniform distribution.
For the development and test sets, this removal procedure was performed when creating the dataset, to improve reproducibility.
For the training set, we retained the original five-sentence story in the dataset and removed sentences randomly when reading the data during training. 
As a result, different sentences could be removed from the same story with different $m$ value, thus, acting as data augmentation and preventing over-fitting.

WritingPrompts is one of the commonly used datasets in the domain of stories \citep{WritingPromptsDataset}. We also considered using this dataset, but ultimately decided not to. The reasons are presented below. 
WritingPrompts consists of 303,358 stories paired with writing prompts collected from an online forum, Reddit.
The average length of the prompt / story is 28.4 / 734.5, respectively. As it contains very long stories, it is generally used with trimming (retain a predetermined number of words from the start and truncate the rest).
In other words, it is difficult to handle the ``whole story'' as is.
We believe that this creates a problem in learning VN-MPP.
This is because VN-MPP is designed to learn by contrasting complete and incomplete texts in order to estimate ``what part of an incomplete text is incomplete.'' Hence, trimmed texts, i.e., texts from which certain parts have been deleted, are not suitable to train VN-MPP models.

\subsubsection{Dataset of Other Domain}

Although we propose VN-MPP mainly for creative support and the target is story-like text, the application of the VN-MPP and proposed methods are not limited only to stories. To show this, we use a dataset from other domains.

\begin{table}[!ht]
\small
\centering
\begin{tabular}{lrr}  
\toprule
set  & \#article and highlights & missing position \\
\midrule
train       & 287,113 & Given when creating dataset \\
dev    & 13,368 & Given when creating dataset \\
test   & 11,490 & Given when creating dataset \\
\midrule
total  & 311,971 & \\
\bottomrule
\end{tabular}
\caption{Overview of the CNN / Daily Mail summarization dataset with our preprocessing.}
\label{tab:dataset_cnn_dailymail}
\end{table}

The CNN / Daily Mail summarization dataset contains online news articles paired with multi-sentence summaries. The average number of tokens of article /summary is 781 / 56, respectively. The original CNN / Daily Mail dataset was proposed to support supervised neural methodologies for machine reading and question answering \citep{Hermann:2015:TMR:2969239.2969428}, and \citet{nallapati-etal-2016-abstractive} modified the dataset to be used for summarization. 
The version we use for our task was proposed by \citet{see-etal-2017-get}. This version of the dataset is non-anonymized; this contrasts with the earlier versions, in which the data are anonymized. They also stated that the non-anonymized version is the favorable problem to solve because it requires no pre-processing.
We also believe that the non-anonymized version is favorable for our task because  it allows the ability to consider proper nouns to be evaluated.

This time, we decided to use highlights instead of articles. This is because ``highlights'' are considered to be more important per sentence, i.e., if they are missing, it would be a big problem.

Moreover, the purpose of using this dataset is to show the versatility of VN-MPP.
In other words, (1) it can be applied to more than just stories (2) it can be applied even if the original text is not a five-sentence text.
For the training set of the CNN / Daily Mail summarization dataset, we examined the mean and standard deviation (std) of the word length and sentence length of the highlights. We found that the mean word length is 54.7 and the std is 23.0. For the number of sentences, the mean is 3.68 and the std is 1.35.
Thus, the highlights of the dataset contain variability and are useful in investigating the adaptability of VN-MPP.

We use the original split as shown in Table \ref{tab:dataset_cnn_dailymail},
We removed sentences from an article. The number of missing positions $m$ was randomly decided to be $ 0 \leq m \leq \min(9, \#sentences) $, based on a discrete uniform distribution.
This removal procedure was performed when creating the dataset, to improve reproducibility.

From the perspective of addressing social media, we considered using the dataset crawled from Twitter, such as Coronavirus (COVID-19) Tweets Datasets (COV19Tweets Dataset) \citep{Lamsal2020}. However, Twitter's terms of service prohibit redistributing tweets as is (because users may want to delete their own tweets). Therefore, the dataset contains the IDs of the tweets rather than the full text of the tweets.
Given Twitter's concern for the rights of users to control their own tweets, we felt it was undesirable to include the original tweets retrieved from Twitter IDs in this paper.
If only automatically generated tweets were included, it would be difficult to make a qualitative evaluation of the proposed task, such as whether completion is indeed achieved.
Although we believe that it is important to handle Twitter, which is a representative example of microblogs, in the study of Social Media, we decided to avoid handling the corpus consisting of Twitter data in this study for this reason.

\subsubsection{Characteristics of the Words in the Two Datasets Used}

To show the trend of word utilization in the dataset we used, we did a word cloud visualization.

\begin{figure}[!ht]
  \center
  \includegraphics[width=\textwidth]{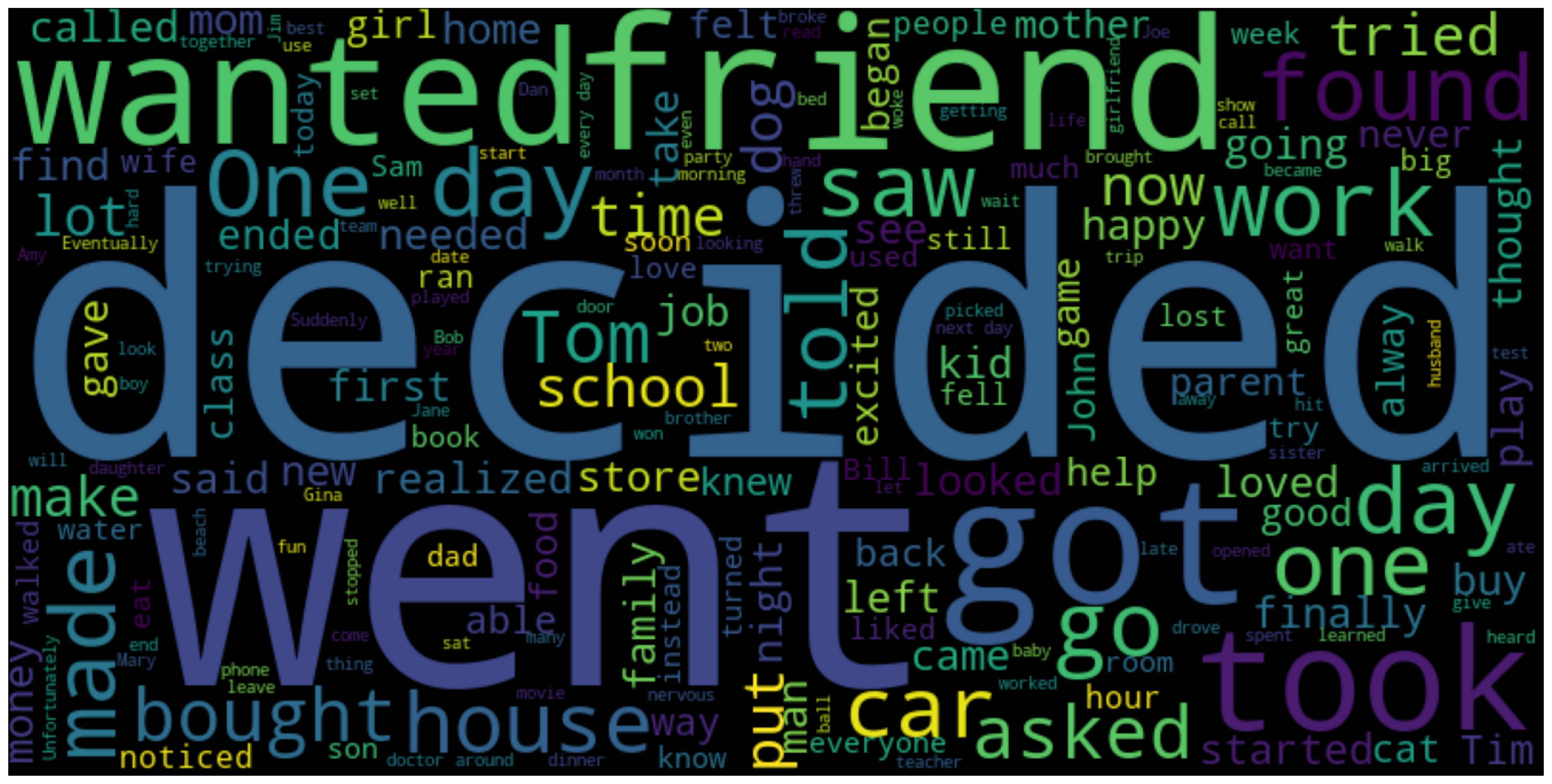}
  \caption{\label{fig:rocstories_word_cloud} Word Cloud of ROCStories.
}
\end{figure}

\begin{figure}[!ht]
  \center
  \includegraphics[width=\textwidth]{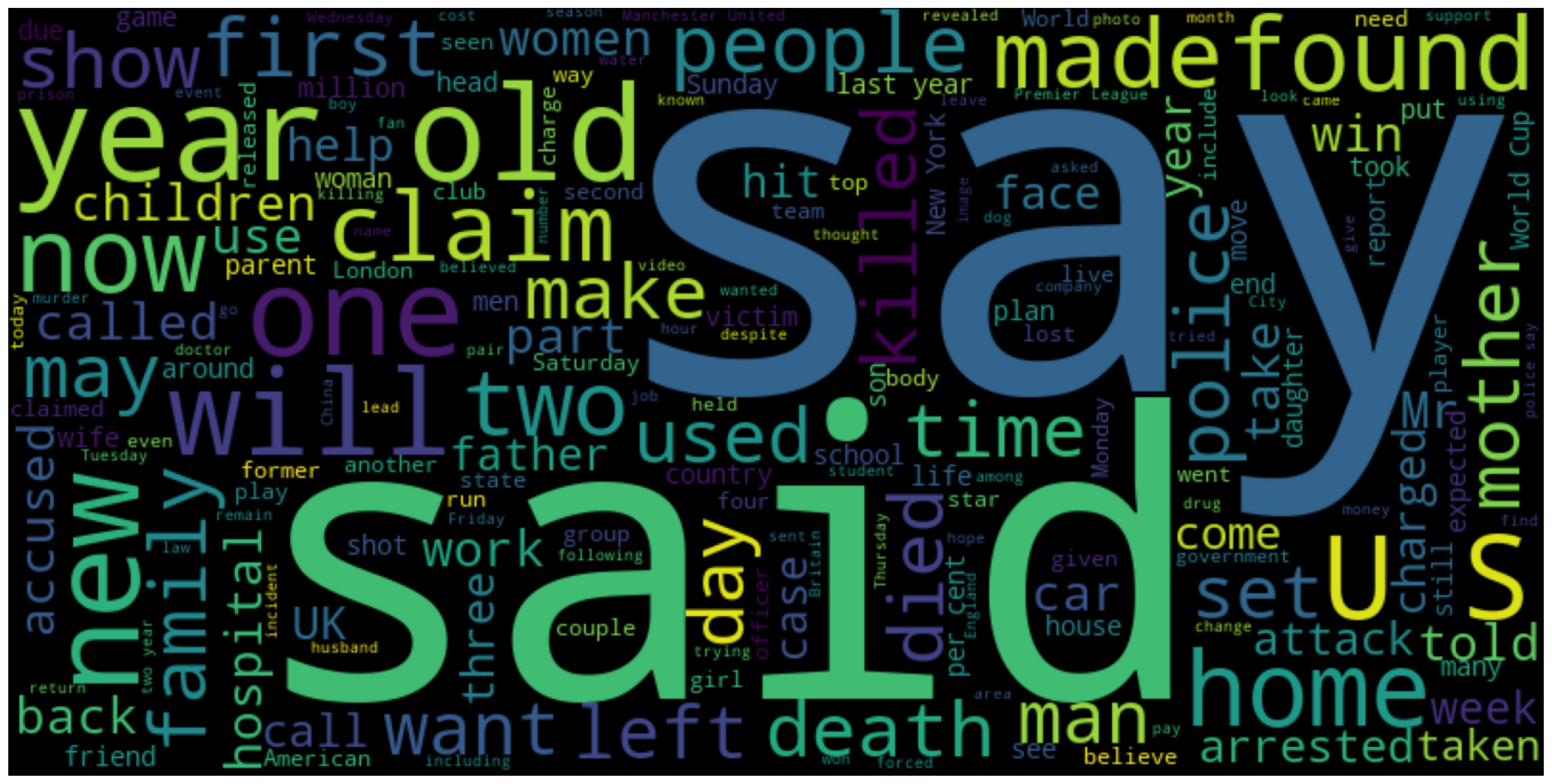}
  \caption{\label{fig:cnndaily_word_cloud} Word Cloud of the highlights in the train set of the CNN / Daily Mail summarization dataset.
}
\end{figure}

Figure \ref{fig:rocstories_word_cloud} shows the Word Cloud of ROCStories, and Figure \ref{fig:cnndaily_word_cloud} shows the Word Cloud of the highlights in the train set of the CNN / Daily Mail summarization dataset.

The former is a dataset of everyday stories, so we can see things like ``went'' and ``friend.'' The latter is a news dataset, so ``say'' and ``said'' stand out, indicating that someone has said something.

We experimented with the concatenation task of VN-MPP and SC.

\subsection{Experiment}

\subsubsection{Methods}
\label{subsec:compare_methods}

The hierarchical method proposed by us \citep{mori-etal-2020-finding} together with their MPP task is based on the limited condition that ``one sentence is missing from a five-sentence text,'' and it is difficult to apply it directly to the VN-MPP task.
VN-MPP is a task that cannot be solved by the conventional methods used for tasks such as MPP, SC, and SEG. In this paper, we propose and evaluate the first two methods for solving VN-MPP.

\subsubsection{Evaluation Metrics}
\label{subsec:eval_metrics}

We first evaluated the proposed method in VN-MPP with each dataset using BLEU, a standard evaluation metric for machine translation \citep{Papineni:2002:BMA:1073083.1073135}.
BLEU is also used in the evaluation of tasks such as SEG, therefore, it can be considered useful in the evaluation of VN-MPP as well.
Furthermore, in VN-MPP using ROCStories, which requires creative text generation, we conducted another evaluation with the recently proposed metrics described in Section \ref{subsection:related_evaluation_metrics}.
Concretely, we used UNION, BERTScore, and BLEURT. 
For UNION, we used the officially provided checkpoint trained with ROCStories.\footnote{\url{https://github.com/thu-coai/UNION}}
For BERTScore and BLEURT, we used their default model ``roberta-large''\footnote{\url{https://github.com/Tiiiger/bert_score}} and ``bleurt-tiny-128''\footnote{\url{https://github.com/google-research/bleurt}}.

\subsubsection{Training Details}
\label{subsec:training_details}

For the two-module approach, we used BART-base or BART-large for each module. When combining two modules, the base models are the same.
For the end-to-end approach, we also used BART-base or BART-large.

BART uses the standard Transformer architecture \citep{Vaswani_2017_transformer}, except that GeLUs~\citep{hendrycks2020gaussian} activation function is used instead of ReLU~\citep{Nair_and_Hinton_2010_ReLU} and parameters are initialised from $\mathcal{N} (0, 0.02)$ \citep{lewis-etal-2020-bart}.
BART-base model has 6 layers for encoder and decoder respectively. BART-large model has 12 layers for each.

We fine-tuned our models over three epochs on NVIDIA Tesla V100 GPUs.
Specifically, we used one GPU for training on ROCStories, and four GPUs for training on the highlights of the CNN / Daily Mail summarization dataset.
We used AdamW \citep{loshchilov2018decoupled} optimization with parameters $\beta_1=0.9, \beta_2=0.999, \epsilon=1e-08$. We set the initial learning rate to $3e-05$, and linearly decreased the learning rate from the initial point to $0$ during 10-epoch training to avoid over-fitting.

Transformer-based Seq2Seq language models have greatly improved performance compared to conventional models in text-to-text tasks, especially in summarization and translation.
To set up the training parameters, we mainly referred to the training settings in the summarization task because we consider summarization is more similar to story completion than translation is.

\subsection{Result}
\label{subsec:result}

\begin{table}[!t]
\small
\centering
\begin{tabular}{lrr}
\toprule
{} &  BLEU &  mean length of text \\
\midrule
VN-MPP module &    96.56 &          45.9 \\
SC module &    80.23 &          52.7 \\
\midrule
End-to-End &    79.71 &          52.9 \\
\bottomrule
\end{tabular}
\caption{BLEU: BART-base on ROCStories.}
\label{tab:result_bart_base_rocstories}
\end{table}

\begin{table}[!t]
\small
\centering
\begin{tabular}{lrr}
\toprule
{} &  BLEU &  mean length of text \\
\midrule
VN-MPP module &    90.92 &          44.1 \\
SC module &    80.31 &          53.1 \\
\midrule
End-to-End &    79.76 &          53.8 \\
\bottomrule
\end{tabular}
\caption{BLEU: BART-large on ROCStories.}
\label{tab:result_bart_large_rocstories}
\end{table}

\begin{table}[!t]
\small
\centering
\begin{tabular}{lrr}
\toprule
{} &  BLEU &  mean length of text \\
\midrule
VN-MPP module &    87.55 &          49.2 \\
SC module &    63.50 &          68.6 \\
\midrule
End-to-End &    54.97 &          47.9 \\
\bottomrule
\end{tabular}
\caption{BLEU: BART-base on CNN / Daily Mail.}
\label{tab:result_bart_base_cnndailymail}
\end{table}

\begin{table}[!t]
\small
\centering
\begin{tabular}{lrr}
\toprule
{} &  BLEU &  mean length of text \\
\midrule
VN-MPP module &    87.18 &          48.9 \\
SC module &    62.30 &          69.3 \\
\midrule
End-to-End &    57.90 &          53.6 \\
\bottomrule
\end{tabular}
\caption{BLEU: BART-large on CNN / Daily Mail.}
\label{tab:result_bart_large_cnndailymail}
\end{table}

\begin{table}[!t]
\small
\centering
\begin{tabular}{lrrr}
\toprule
{} &  UNION & BERTScore & BLEURT \\
\midrule
Two-module (base) &    0.911 & 0.974 & 0.452 \\
End-to-End (base) &    0.905 & 0.974 & 0.446 \\
\cmidrule(lr){1-4}
Two-module (large) &   0.806 & 0.972 & 0.469 \\
End-to-End (large) &   0.904 & 0.974 & 0.439 \\

\bottomrule
\end{tabular}
\caption{Other Metrics: BART-base (shown as base) and BART-large (large) on ROCStories.}
\label{tab:result_roc_other_metrics}
\end{table}

\begin{table}[!t]
\footnotesize
\center
\begin{tabular}{p{2.5cm}p{13cm}}
\toprule
{} &                                                                                                                                                                                                     Example 0 \\
\midrule
context                 &                                                                    He went to the dealership and bought a really fancy BMW. He showed it off around town. Evan knew he looked cool in the new car. \\
\cmidrule(lr){1-2}
missing\_ids             &                                                                                                                                                                                             [0, 2] \\
target\_sentences        &                                                                                                                               [Evan had been saving for years., Evan was so proud of his new car.] \\
original\_story          &  Evan had been saving for years. He went to the dealership and bought a really fancy BMW. Evan was so proud of his new car. He showed it off around town. Evan knew he looked cool in the new car. \\
\cmidrule(lr){1-2}
VN-MPP module output          &                              <missing\_sentence> He went to the dealership and bought a really fancy BMW. He showed it off around town. <missing\_sentence> Evan knew he looked cool in the new car. \\
SC module output          &    Evan wanted to buy a new car. He went to the dealership and bought a really fancy BMW. He showed it off around town. He drove it all the way home. Evan knew he looked cool in the new vehicle. \\
\cmidrule(lr){1-2}
End-to-End & Evan wanted to buy a new car. He went to the dealership and bought a really fancy BMW. He showed it off around town. He drove it home and had a great time. Evan knew he looked cool in the new car! \\
\midrule
{} &                                                                                                                                                                                                                                                                                    Example 1 \\
\midrule
context                 &                                                     Serena was planning a surprise for her husband's birthday. She wanted to throw him a party, but his schedule was tough. He would always arrive home at widely different times. To get around it, she worked with his co-workers. \\
\cmidrule(lr){1-2}
missing\_ids             &                                                                                                                                                                                                                                                                                  [4] \\
target\_sentences        &                                                                                                                                                                                                                                 [Together they were able to surprise him perfectly.] \\
original\_story          &  Serena was planning a surprise for her husband's birthday. She wanted to throw him a party, but his schedule was tough. He would always arrive home at widely different times. To get around it, she worked with his co-workers. Together they were able to surprise him perfectly. \\
\cmidrule(lr){1-2}
VN-MPP module output          &                                  Serena was planning a surprise for her husband's birthday. She wanted to throw him a party, but his schedule was tough. He would always arrive home at widely different times. To get around it, she worked with his co-workers. <missing\_sentence> \\
SC module output          &     Serena was planning a surprise for her husband's birthday. She wanted to throw him a party, but his schedule was tough. He would always arrive home at widely different times. To get around it, she worked with his co-workers. She was able to throw a surprise party for him. \\
\cmidrule(lr){1-2}
End-to-End & Serena was planning a surprise for her husband's birthday. She wanted to throw him a party, but his schedule was tough. He would always arrive home at widely different times. To get around it, she worked with his co-workers. She was able to throw a surprise party for him. \\
\bottomrule
\end{tabular}
\caption{Generation results of BART-base based fine-tuned models on ROCStories.}
\label{tab:result_bart_base_generation_rocstories}
\end{table}

\begin{table}[!t]
\footnotesize
\center
\begin{tabular}{p{2.5cm}p{13cm}}

\toprule
{} &                                                                                                                                                                                                     Example 0 \\
\midrule
context                 & "Hazzard" ran from 1979 to 1985 and was among the most popular shows on TV .                                                                       \\
original highlights          & James Best, who played the sheriff on "The Dukes of Hazzard," died Monday at 88 .\verb|\n|"Hazzard" ran from 1979 to 1985 and was among the most popular shows on TV .    \\
\cmidrule(lr){1-2}
VN-MPP module output          & "Hazzard" ran from 1979 to 1985 and was among the most popular shows on TV. \verb|<missing_sentence>| \\
SC module output          &  "Hazzard" ran from 1979 to 1985 and was among the most popular shows on TV.\verb|\n|The show was canceled in 2008.\\
\cmidrule(lr){1-2}
End-to-End & "Hazzard" ran from 1979 to 1985 and was among the most popular shows on TV. \\
\\
\midrule
{} &                                                                                                                                                                                                     Example 1 \\
\midrule
context                 & Moschetto, 54, was arrested for selling drugs and weapons, prosecutors say .                                                                       \\
original highlights          &  A lawyer for Dr. Anthony Moschetto says the charges against him are baseless .\verb|\n|Moschetto, 54, was arrested for selling drugs and weapons, prosecutors say .\verb|\n|Authorities allege Moschetto hired accomplices to burn down the practice of former associate .   \\
\cmidrule(lr){1-2}
VN-MPP module output          & \verb|<missing_sentence>| Moschetto, 54, was arrested for selling drugs and weapons, prosecutors say. \\
SC module output          &  NEW: Prosecutor: "This is not the first time we've seen this type of behavior"\verb|\n|NEW: Moschetto's attorney says he will appeal the decision.\verb|\n|Moschetto, 54, was arrested for selling drugs and weapons, prosecutors say.\\
\cmidrule(lr){1-2}
End-to-End & Michael Moschetto is accused of killing his wife and two children.\verb|\n|Moschetto, 54, was arrested for selling drugs and weapons, prosecutors say. \\
\bottomrule
\end{tabular}
\caption{Generation results of BART-base based fine-tuned models on CNN / Daily Mail.}
\label{tab:result_bart_base_generation_cnndaily}
\end{table}

The results of the evaluation with BLEU are shown in Tables \ref{tab:result_bart_base_rocstories}, \ref{tab:result_bart_large_rocstories}, \ref{tab:result_bart_base_cnndailymail}, and \ref{tab:result_bart_large_cnndailymail}. 
The average length of the generated text is also shown.
Note that the two-module approach evaluated the VN-MPP module and SC module separately, using the input and output of each module.

The results for the BART-base based models on ROCStories are presented in Table \ref{tab:result_bart_base_rocstories}, and those for the BART-large based models on ROCStories in Table \ref{tab:result_bart_large_rocstories}.
In the same manner, the results for the models based on BART-base or BART-large and fine-tuned on the CNN / Daily Mail summarization dataset are shown in Table \ref{tab:result_bart_base_cnndailymail} and Table \ref{tab:result_bart_large_cnndailymail}, respectively.

For the VN-MPP module, good results are obtained especially for ROCStories, and high BLEU values are obtained for the CNN / Daily Mail summarization dataset. 
Although not as good as the VN-MPP module, the SC module and the End-to-End module also show sufficient performance considering that they include sentence generation in their tasks.

Table \ref{tab:result_roc_other_metrics} shows the results of evaluation with other metrics.
Note that for these metrics the two-module approach was evaluated as a whole, not separately. 
The reason is that UNION, in particular, is a measure for evaluating whether a text is story-like or not, and is inappropriate for evaluating the VN-MPP module, which outputs a text that includes a special token.
For the same reason, we do not use these metrics to evaluate CNN / Daily Mail.
UNION is a binary classification model; hence, we show the ratio of the number of output sentences judged to be ``story-like'' in the test set (9817 instances), with a probability of 0.5 as a threshold.

Only the two-module approach using BART-large has a slightly low UNION value; the other models have values above 0.9, indicating that story-like outputs can be generated (completed) by our methods. The values for BERTScore and BLEURT are close.

Examples of the results of VN-MPP + SC solved by our methods on ROCStories and CNN / Daily Mail are shown in Tables \ref{tab:result_bart_base_generation_rocstories} and \ref{tab:result_bart_base_generation_cnndaily}, respectively.
Regarding the experimental results obtained using ROCStories, it can be seen that the number and position of the missing sentences can be correctly predicted even when the number of missing sentences differs. Furthermore, in terms of meaning, sentences that are natural in the context of the story can be generated.
On the other hand, on CNN / Daily Mail, the model failed to estimate the missing positions, or generated sentences that were close in mood but not necessarily in context.

\section{Creative Support System with VN-MPP}
\label{sec:demonstration-system}

Based on the task and methods proposed in Sections \ref{sec:task_description} and \ref{sec:methodology}, we built a system that executes VN-MPP and SC. We named this system \textbf{``COMPASS''}, which stands for a writing support system to \textbf{COMP}lement \textbf{A}uthor unaware \textbf{S}tory gap\textbf{S}.
As explained above, VN-MPP excludes the constraints of LC-MPP and allows the number of sentences in the input to vary. More specifically, the number of sentences in the original story and the number of missing sentences can vary.

\begin{figure}[!t]
    \begin{center}
    \includegraphics[clip,width=\linewidth]{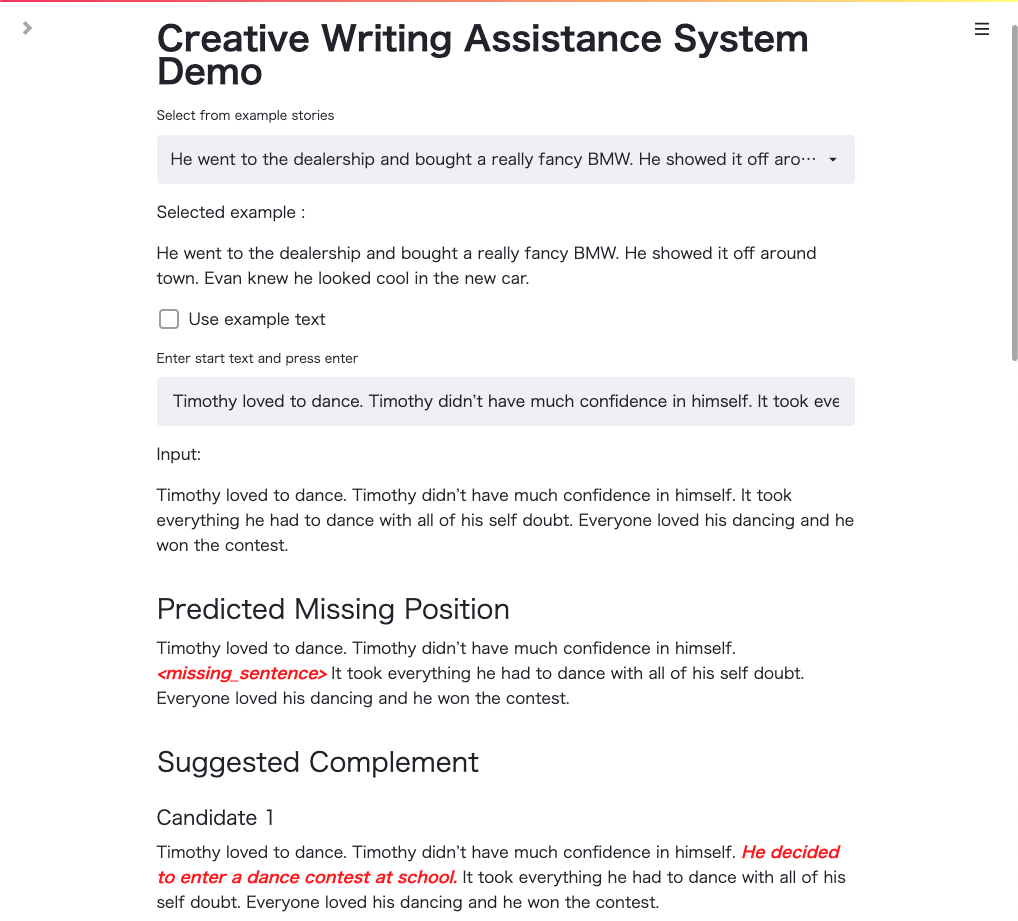}
    \end{center}
    \caption{The VN-MPP + SC demo system for human story writing assistance. It estimates the missing positions of a given incomplete story and generates and presents sentences to complete the story.}
    \label{fig:vn-mpp_demo}
\end{figure}

Figure \ref{fig:vn-mpp_demo} shows the appearance of this system. 

Moreover, we introduce beam search as a decoding algorithm for the system instead of greedy sampling.\footnote{Greedy sampling is also known as greedy search or the greedy algorithm.}
As can be seen in \citep{Bahdanau2014,freitag-al-onaizan-2017-beam}, beam search was already being used in Seq2seq models in the early days when Seq2seq models using RNN were proposed \citep{Graves2012,DBLP:conf/ismir/Boulanger-LewandowskiBV13,Sutskever:2014:SSL:2969033.2969173}.
By using beam search, multiple candidate sentences can be displayed. 
We made the beam size and the number of suggested sentences user-selectable, so that users can receive more interactive assistance with story completion.

\subsection{Functions and Usability}

Based on the user input, our system can identify the parts to be completed in incomplete stories and generate candidate sentences for completion.
In addition to inputting arbitrary sentences, the user can also use pre-prepared example sentences, which makes it easy to test the system's usability.

In the prototype system we have already made publicly available,\footnote{\url{https://github.com/mil-tokyo/mppsc-demo}} the input is limited to ``four sentences with one sentence missing from a story consisting of five sentences."
Furthermore, only one sentence is generated as a candidate sentence.

The system using VN-MPP, which is an extension of LC-MPP, removes these limitations and can estimate multiple missing locations for variable-length inputs.
In addition, we developed a new two-module v2 system that enables us to present multiple candidates by beam search.

We also added functions to display some information that may be useful to users.
One such function evaluates the story-likeness of the completed text. For this function, we use UNION as a metric \citep{UNION}. Note that we trained our own model of UNION using DistilBERT \citep{Sanh2019DistilBERTAD} instead of BERT, which was used in the original implementation. 
Another function visualizes the emotions of the reader.
In our previous study~\citep{MORI20191865}, we proposed ``Emotional Flow'' and showed the importance of emotions, especially multi-perspective and multi-dimensional emotions.
Considering emotions is a well-known approach in storytelling; hence, we include emotion visualization as an essential part of our system.
To predict emotions from input and output text, we fine-tuned BERT with ``EmoBank'' 
EmoBank is proposed as a text corpus manually annotated with emotion according to the psychological Valence-Arousal-Dominance (VAD) scheme \citep{buechel-hahn-2017-emobank,buechel-hahn-2017-emobank-readers-vs-writers-vs-texts}.
This dataset contains annotations from two perspective: the reader perspective and the writer perspective. We focus on the annotation from the reader perspective. As our intention is to focus on stories, we decided to select ``fiction'' and ``essays'' from all categories, considering the characteristics of each category.
Using the predicted Valence and Arousal value, we draw Emotional Flow.

\subsection{Methodology: updated method for Creative Support System}
\label{subsec:updated_methodology}

Although the end-to-end method is considered the mainstream in the field of machine learning, in the first method, we deliberately divided the module for each task so that the output of VN-MPP can be confirmed independently.
In addition, because various methods for SC have been proposed and are expected to be developed in the future, it is also advantageous that our proposed VN-MPP module can be easily combined with these methods.

\begin{figure}[!t]
  \center
  \includegraphics[width=\textwidth]{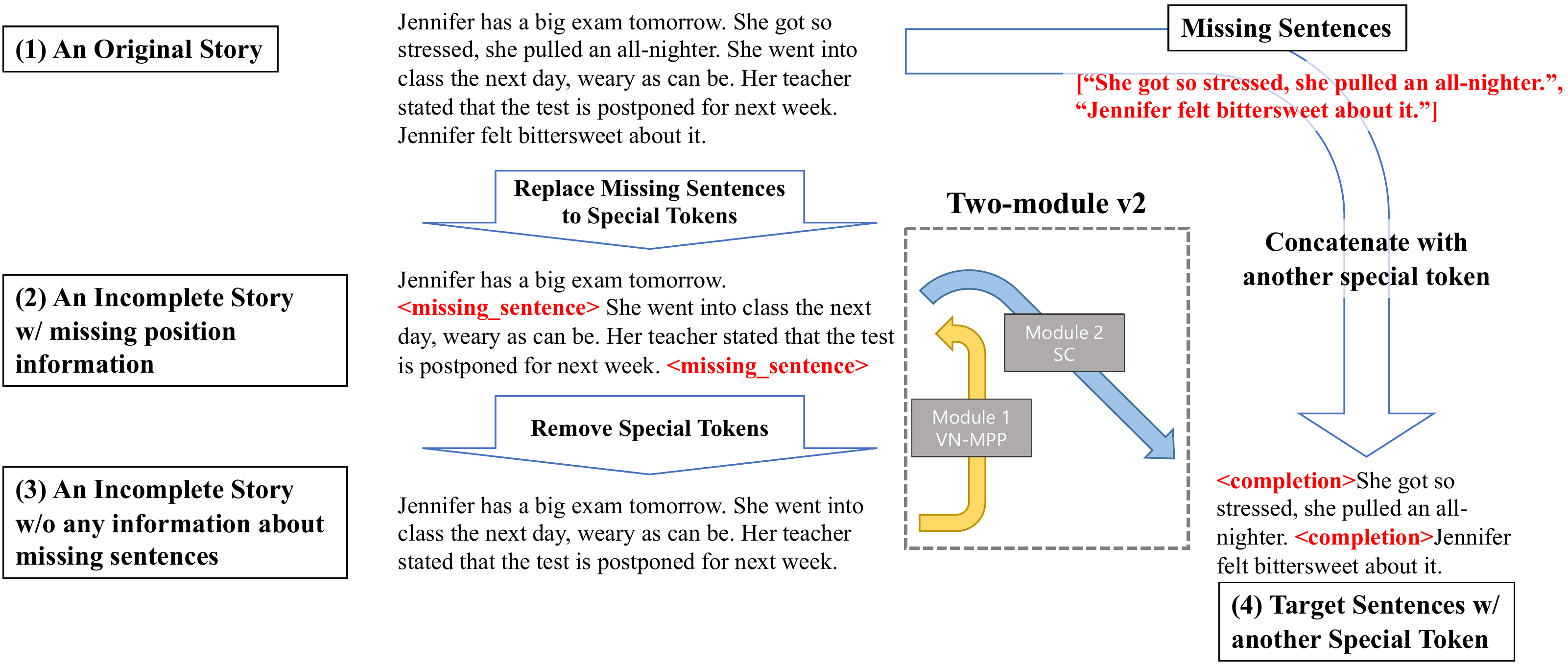}
  \caption{\label{fig:two-module_v2_preprocessing} Preprocessing the dataset in two-module v2 and correspondence between the input and output of each module.
}
\end{figure}

As a comparison with Figure \ref{fig:dataset_preprocessing}, we show the new dataset preprocessing method and the input/output of the two modules in Figure \ref{fig:two-module_v2_preprocessing}.

As before, the same process is used to create a sentence in which a part of the input sentence is replaced by \verb|<missing_sentence>| and an incomplete story that does not contain any information about the missing position. However, it additionally uses the original sentence that was replaced by \verb|<missing_sentence>|, i.e., missing sentences.

The problem is that the number of missing sentences is variable in VN-MPP and, assuming that they are generated separately, the number of times the output is performed depends on the input, which makes batch learning difficult.

Therefore, we conceived the idea of introducing yet another special token.
Specifically, we added another special token, \verb|<completion>|, to the beginning of each missing sentence, so that the missing sentence can be treated as a single sequence. Then, after generating the missing sentences, we use the \verb|<completion>| token as a guide to retrieve the complement sentences from the generated sequence.
This allows the new Module 2 to generate only completion statements, and to control them without affecting the rest of the system.
\footnote{The implementation of this method occurred with the mistyping of ``\textless completion\textgreater''~as ``\textless complition\textgreater.'' However, this token is registered in the tokenizer independently of the original word (completion), and is not affected by the string in the token. Additionally, the proposed system does not display the token, so users would not see the typo in the token. Therefore, this typo does not affect the evaluation experiment, nor does it affect the claims in this thesis.}

\subsection{Implementation Details}
\label{subsec:implementation-details_system}

The front-end and back-end of this web application are written in Python using Streamlit.\footnote{\url{https://streamlit.io/}}
For machine learning model implementation, we mainly use PyTorch and HuggingFace Transformers.

\section{User Study with Japanese-version System}
\label{sec:vnmpp-user-study}

To verify the usefulness of our proposed system from the viewpoint of creative writing assistance, we conducted a user study.
Specifically, we asked professional storytellers in Japan to evaluate it.

\subsection{Overview of the Japanese-version System}
\label{subsec:ja-system}

\begin{figure}[!pt]
    \begin{center}
    \includegraphics[clip,width=\linewidth]{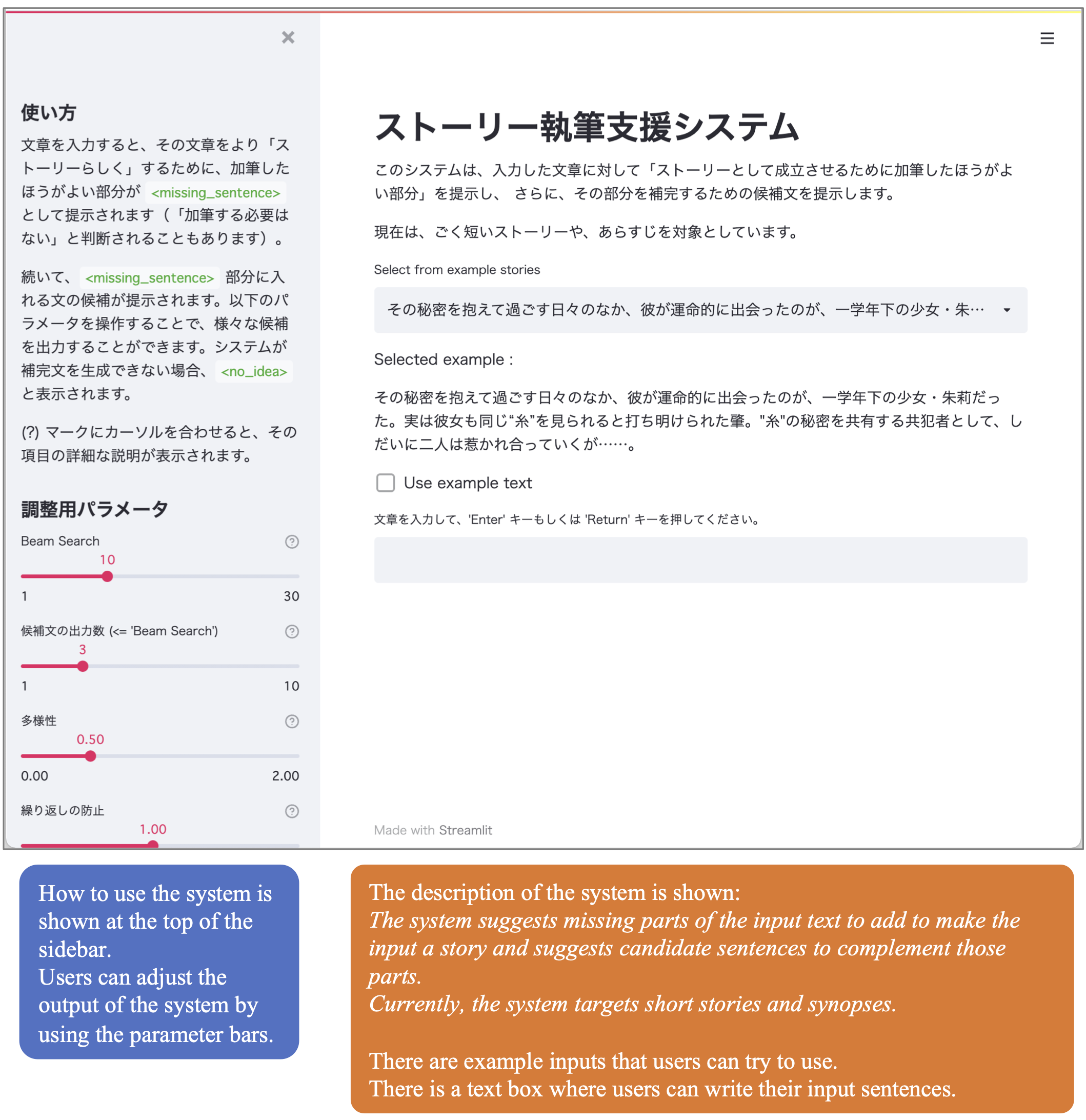}
    \end{center}
    \caption{Initial state of the Japanese version of the proposed system.}
    \label{fig:ja-system_first_look}
\end{figure}

\begin{figure}[!pt]
    \begin{center}
    \includegraphics[clip,width=\linewidth]{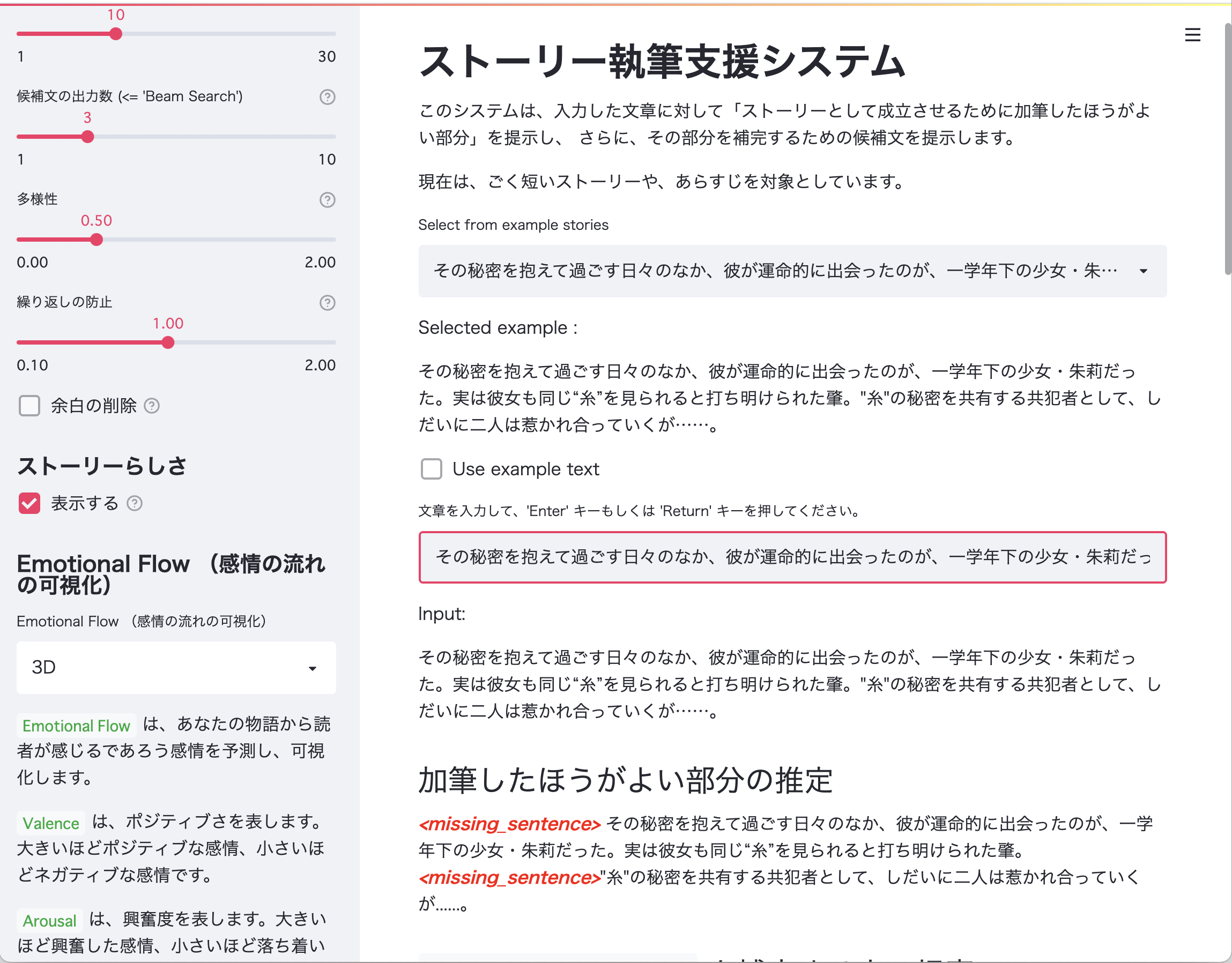}
    \end{center}
    \caption{The user inputs an arbitrary text and presses the ``Enter'' or the ``Return'' key, then our VN-MPP is executed. The input is repeated and the result of VN-MPP is displayed. In the example in this figure, there are two missing position predicted.}
    \label{fig:ja-system_mpp}
\end{figure}

\begin{figure}[!pt]
    \begin{center}
    \includegraphics[clip,width=\linewidth]{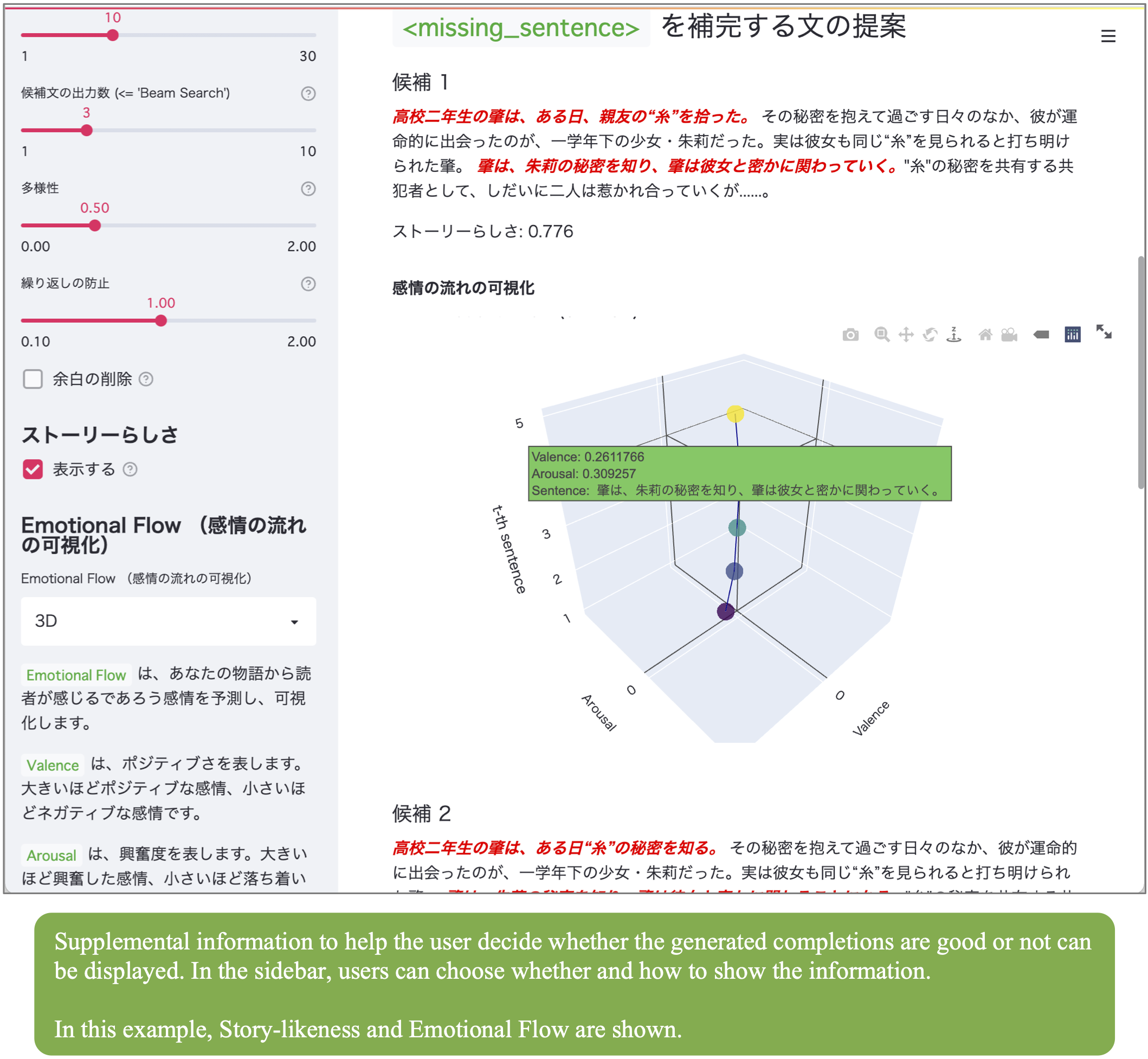}
    \end{center}
    \caption{Candidate sentences to insert into the missing positions are presented. Emotional Flow is also displayed, allowing the user to see the kinds of emotions each sentence will evoke in the reader.}
    \label{fig:ja-system_canditate_and_emotional_flow}
\end{figure}

\begin{figure}[!pt]
    \begin{center}
    \includegraphics[clip,width=\linewidth]{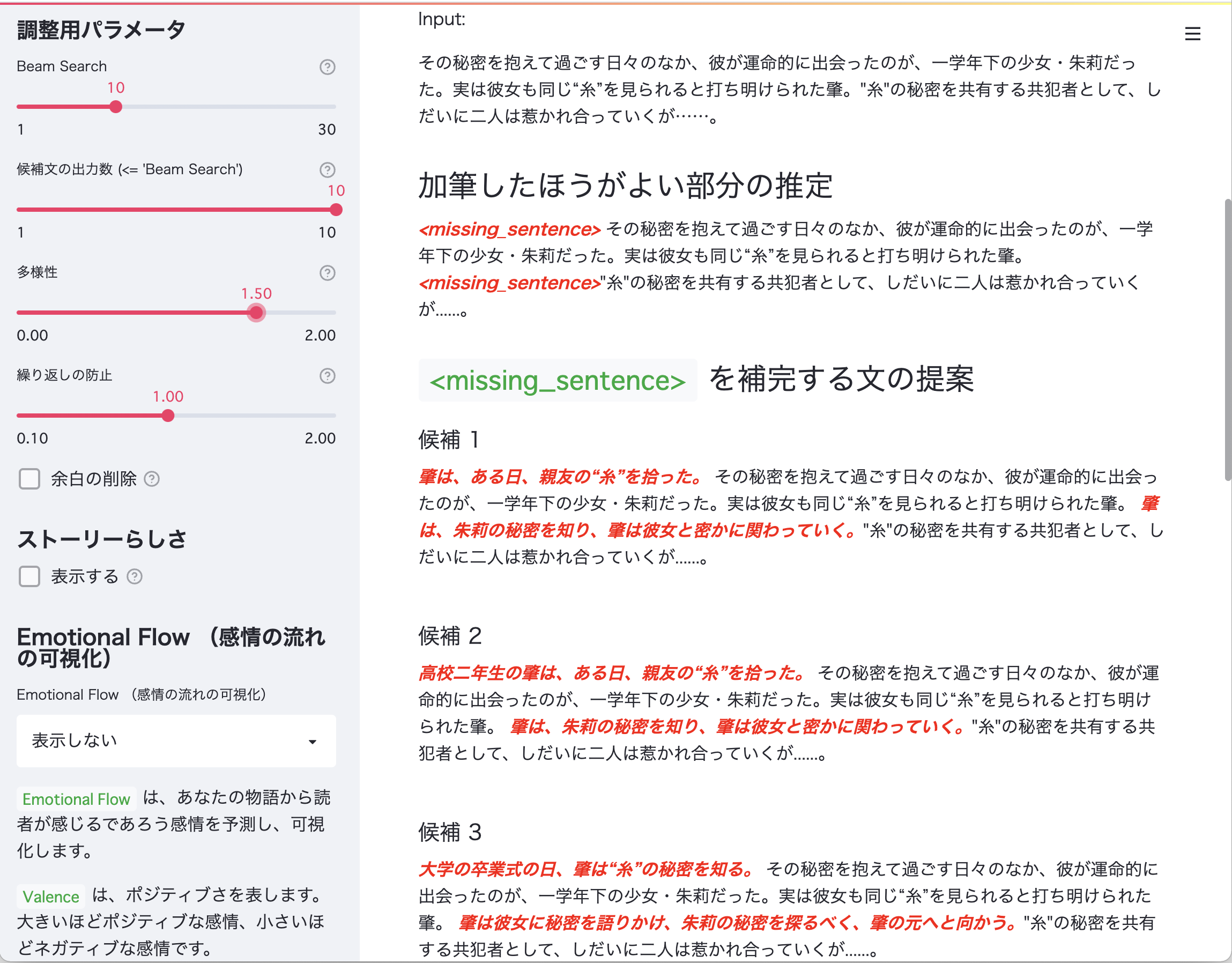}
    \end{center}
    \caption{The user can adjust the parameters to find desirable candidate sentences. In Candidate 2, the protagonist, ``肇 (Hajime),'' is said to be a ``高校二年生 (sophomore in high school)'' (which matches the setting of the original work used for the input sentence), and in Candidate 3, the setting of a college student is suggested because it is a ``大学の卒業式 (college graduation ceremony).''}
    \label{fig:ja-system_candidate_completion_sentences}
\end{figure}

When a user accesses the system, the screen shown in Figure \ref{fig:ja-system_first_look} is presented.
Then, the user enters an arbitrary sentence, and the system returns the missing position, as shown in Figure \ref{fig:ja-system_mpp}.
In addition, candidate sentences to insert into the missing positions are presented (Figure \ref{fig:ja-system_canditate_and_emotional_flow}).
Emotional Flow is also displayed, allowing the user to see the kinds of emotions the sentence will evoke in the reader.
The user can adjust the parameters to find desirable candidate sentences while considering the emotion the user wants to evoke in the reader (Figure \ref{fig:ja-system_candidate_completion_sentences}).

The example input used in the series of images and its English translation are as follows:
\begin{itemize}
    \item その秘密を抱えて過ごす日々のなか、彼が運命的に出会ったのが、一学年下の少女・朱莉だった。実は彼女も同じ糸を見られると打ち明けられた肇。“糸”の秘密をめぐる共犯者として、しだいに二人は惹かれ合っていくが……。
    \item (English translated version) While spending his days with this secret, he fatefully meets a girl one year younger than him, Akari. Hajime is told that she can also see the same strings. As accomplices in the secret of the ``strings,'' they gradually become attracted to each other.
\end{itemize}

The example input is a modified version of the synopsis of the following work: 『僕らふたりに運命の糸は』（霧友正規, KADOKAWA 富士見Ｌ文庫, 2019）.\footnote{\url{https://lbunko.kadokawa.co.jp/product/kiritomo/321908000504.html}}
We artificially made the following information missing: Hajime is a high school student who can see the ``red thread of fate'' of anyone he touches.
As shown in Figure \ref{fig:ja-system_candidate_completion_sentences}, our system can generate the concrete content equivalent of ``he'' and ``this secret'' before the pronoun and the demonstrative adjective appear.

\subsection{Development of the Japanese-version System for User Study}
\label{subsec:develop_japanese_system}

\subsubsection{Pre-trained Model Used for Fine-tuning}

To handle Japanese input, we use mBART50, a multilingual Sequence-to-Sequence model proposed by \citet{tang2020multilingual}.
They state that mBART \citep{liu-etal-2020-multilingual-denoising}, referred to as an example of previous multilingual models, has been trained on a variety of languages, but the multilingual nature of the pre-training is not used during finetuning. 
They proposed ``multilingual finetuning'' as a replacement for bilingual finetuning and demonstrated large improvements.

We use the pre-trained checkpoint ``facebook/mbart-large-50'' for finetuning on the Japanese-version tasks: VN-MPP and SC.\footnote{\url{https://huggingface.co/facebook/mbart-large-50}}

\subsubsection{Dataset Preparation}
\label{subsec:japanese_dataset}

To develop the Japanese version of our system, we needed a dataset of stories written in Japanese.
We considered the following two approaches.

\begin{itemize}
    \item Gather stories written in Japanese and use  them as the Japanese dataset.
    \item Translate the English dataset to Japanese and use it as the Japanese dataset.
\end{itemize}

We chose the former approach, and constructed a Japanese novel synopsis dataset: \textbf{Narou-synopsis}.
\textbf{``Shosetsuka ni Narou (小説家になろう)''} is a web service that allows users to post their original novels. Users can post and read novels for free.
The name of the service is a registered trademark of HinaProject Inc. and the company provides the API to obtain the metadata of the novels posted to the website.\footnote{\url{https://dev.syosetu.com/man/api/}} 
We use this API to retrieve the metadata of the novels. 
The available metadata include a synopsis, which we utilize as training data.

We retrieved metadata for a total of 40519 novels from 21 genres. The service assigns points to novels based on readers' responses, and we obtained up to 2,000 entries in each genre, in order of the highest total points. There were some genres with less than 2,000 points, hence the total number was less than 42,000. 

As a second approach, we first attempted to add a translation module to the English version of the system that we developed earlier.
Specifically, for each Japanese input, we obtained an English translation of the input sentence by means of a Japanese--English translation system. The system then evaluated it, and subsequently performed back translation between English and Japanese to obtain the Japanese output.
However, we deemed this method unsuitable for novel texts because the back translation changes the style and nuance of the original text.

Therefore, we attempted to use machine translation before learning; specifically, we used all the ROCStories that had been machine translated into Japanese as the training data.
We named this dataset \textbf{ROCStories-auto-Ja}. 

For machine translation, we used mBART50 finetuned for multilingual machine translation, named ``facebook/mbart-large-50-many-to-many-mmt.''\footnote{\url{https://huggingface.co/facebook/mbart-large-50-one-to-many-mmt}}

\subsubsection{Concatenation of the VN-MPP module and the SC module}
\label{subsec:concat_module1_and_2}

In this multilingual scenario, we discovered an advantage of the two-module approach: the VN-MPP module and the SC module can be trained on different datasets and then combined.

The system using ROCStories-auto-Ja showed good performance in MPP. However, because the complement sentences it generated were not originally written in Japanese, it was difficult to capture the context of proper nouns.

Although the MPP validation score of Narou-synopsis was lower than that of ROCStories-auto-Ja, the SC module was able to generate sentences that are typical in modern Japanese entertainment novels for young people.

Therefore, we used a system that combines the VN-MPP module using ROCStories-auto-Ja and the SC module using Narou-synopsis for the user study.

The model used for the translation was ``facebook/mbart-large-50-many-to-many-mmt,'' the same checkpoint of mBART50 that we used for creating ROCStories-auto-Ja.

\subsection{User Study Design}

We asked four professionals in the creative writing field to evaluate our proposed system. 
We prepared three systems: our proposed system and two comparison systems. The comparison systems were designed and implemented without MPP ability.

\begin{itemize}
    \item System A: \textbf{COMPASS} (Proposed System)
    \item System B: ``Always Add Last'' System
    \item System C: ``Random MPP'' System
\end{itemize}

The order of the three systems was determined randomly.\footnote{The random seed was set to 42, and the list of systems was sorted by NumPy.} All three systems were identical in appearance, except for the alphabet assigned to the system.

``Always Add Last'' is a system that always determines that the end is missing.
This comparison system was designed with reference to \textbf{Write With Transformers},\footnote{\url{https://transformer.huggingface.co/}} which are text auto-completion systems in which Causal Language Models such as GPT-2 are used for generating subsequent text. Moreover, the appearance of the system was aligned for a fair comparison with our proposed system.

In ``Random MPP,'' a randomly selected position was presented as the ``missing position'' instead of the output of Module 1 (VN-MPP). Random seeds were set to return the same result for the same input. 

The design of the ``Random MPP'' system was the most difficult part. A system that ``points out different positions as missing for the same input each time it is executed'' could be considered.
This is the equivalent of advice from someone who changes their opinion from time to time.
However, as our intention was not to frustrate the user, we designed ``Random MPP'' as described above.

As the computing environment for executing the systems, we used Amazon Elastic Compute Cloud (Amazon EC2) of Amazon Web Services (AWS).\footnote{\url{https://aws.amazon.com/ec2/}} 
To ensure fairness, we executed the three systems on equivalent virtual servers.
On Amazon EC2, we chose the ``g4dn.xlarge'' instance for the virtual server.

We had the evaluators view an instructional video explaining how to use the systems.
For this instructional video, we used an earlier version of the proposed system instead of the version used for the evaluation, i.e., System A. This is because it could be assumed to be the proposed system when there was a match with the example in the instructional video.
Then, we had the evaluators access and use the three systems and fill out a questionnaire with their ratings.

This experiment was designed to allow the evaluators to participate remotely.

\subsection{Questionnaire}

The questions we asked the users to answer are shown in Tables \ref{table:user_study_questionnaire_1} and \ref{table:user_study_questionnaire_2}. 
The questions in Table \ref{table:user_study_questionnaire_1} constituted the first half of the questionnaire, and involved user evaluation and comparison of the systems.
The questions in Table \ref{table:user_study_questionnaire_2} constituted the remaining part, and sought to determine how the users create stories and what they desire from creative support systems.
Note that the original questionnaire we used was written in Japanese. 

\begin{table}[p!]
  \centering
  \small
  \begin{tabular}[t]{p{6cm}p{5.5cm}c}
    \toprule
    Question & Choices & Required  \\
    \midrule
    \multicolumn{3}{c}{Section 1 - Compare among systems} \\
    \midrule
    Select the systems in order of preference. & A, B, C & * \\
    Write the reason of your answer in the previous question. & Free writing & * \\
    \midrule
    \multicolumn{3}{c}{Section 2, 3, 4 - Evaluate system A, B, C for each} \\
    \midrule
    Evaluate the prediction of where to add. & 5-point scale & * \\
    Write the reason for your answer to the previous question. & Freewriting  &  \\
    Evaluate the generation of complementary sentence. & 5-point scale & * \\
    Write the reason for your answer to the previous question. & Freewriting & \\
    Select functions you think useful for story creation (as many as you like). & \begin{tabular}{p{5cm}}
        - Predict where to add \\ - Propose complementary sentences \\ - Parameters to adjust output \\ - Predict Story-likeness \\ - Predict and visualize the reader emotions \\ - Others 
    \end{tabular} & * \\
    \\
    Write the reason for your answer to the previous question. & Freewriting & \\
    If you have created a story using this system that you think is a good story and you think it would be OK to publish it, write it in the form. & Freewriting & \\
    \bottomrule
  \end{tabular}
  \caption{First half of the questionnaire used in the user study. With ``Others,'' a free description box was provided. Regarding the question about useful function, if the user felt that there was nothing useful, we asked that they select ``Other'' and indicate that. ``*'' in the ``Required'' row indicates that the answer is required.}
  \label{table:user_study_questionnaire_1}
\end{table}

\begin{table}[p!]
  \centering
  \small
  \begin{tabular}[t]{p{6cm}p{5.5cm}c}
    \toprule
    Question & Choices & Required  \\
    \midrule
    \multicolumn{3}{c}{Section 5 - How you create stories} \\
    \midrule
    Are you (or, were you) a professional storyteller? & Yes / No / No Answer / Others & * \\
    What process do you use to create a story? & 
    \begin{tabular}{p{5cm}}
        - Think in order from the beginning \\ - Write from the place you can think of and try filling in the gaps. \\ - Others 
    \end{tabular} & \\ 
    \\
    Are there any theories or books that you refer to in your creative process? & 
    \begin{tabular}{p{5cm}}
        - Hero's Journey \\ - Three-Act Structure. \\ - Blake Snyder Beat Sheet \\ - Emotional Arc \\ - Others 
    \end{tabular} & \\ 
    \midrule
    \multicolumn{3}{c}{Section 6 - What do you need for Creative Support System} \\
    \midrule
    What do you think is necessary and what would you like to see in a creative support system? & Freewriting & \\
    \bottomrule
  \end{tabular}
  \caption{Second half of the questionnaire used in the user study.}
  \label{table:user_study_questionnaire_2}
\end{table}

\subsection{Results and Discussion}
\label{subsec:user-study-result}

Here, we discuss the evaluation results obtained from the user study.

All comments in the free writing field were written in Japanese. 
We have included the original responses by the evaluators as they were written.
The English translations are given by us for reference purposes.

\subsubsection{Ranking the Systems}
\label{subsubsec:ranking_systems}

In the questionnaire, we first ask each evaluator to individually rank the systems as best, second best, and worst based on their opinion. We also ask each of them to write the reason for their ranking.

\begin{figure}[!t]
    \begin{center}
    \includegraphics[clip,width=\columnwidth]{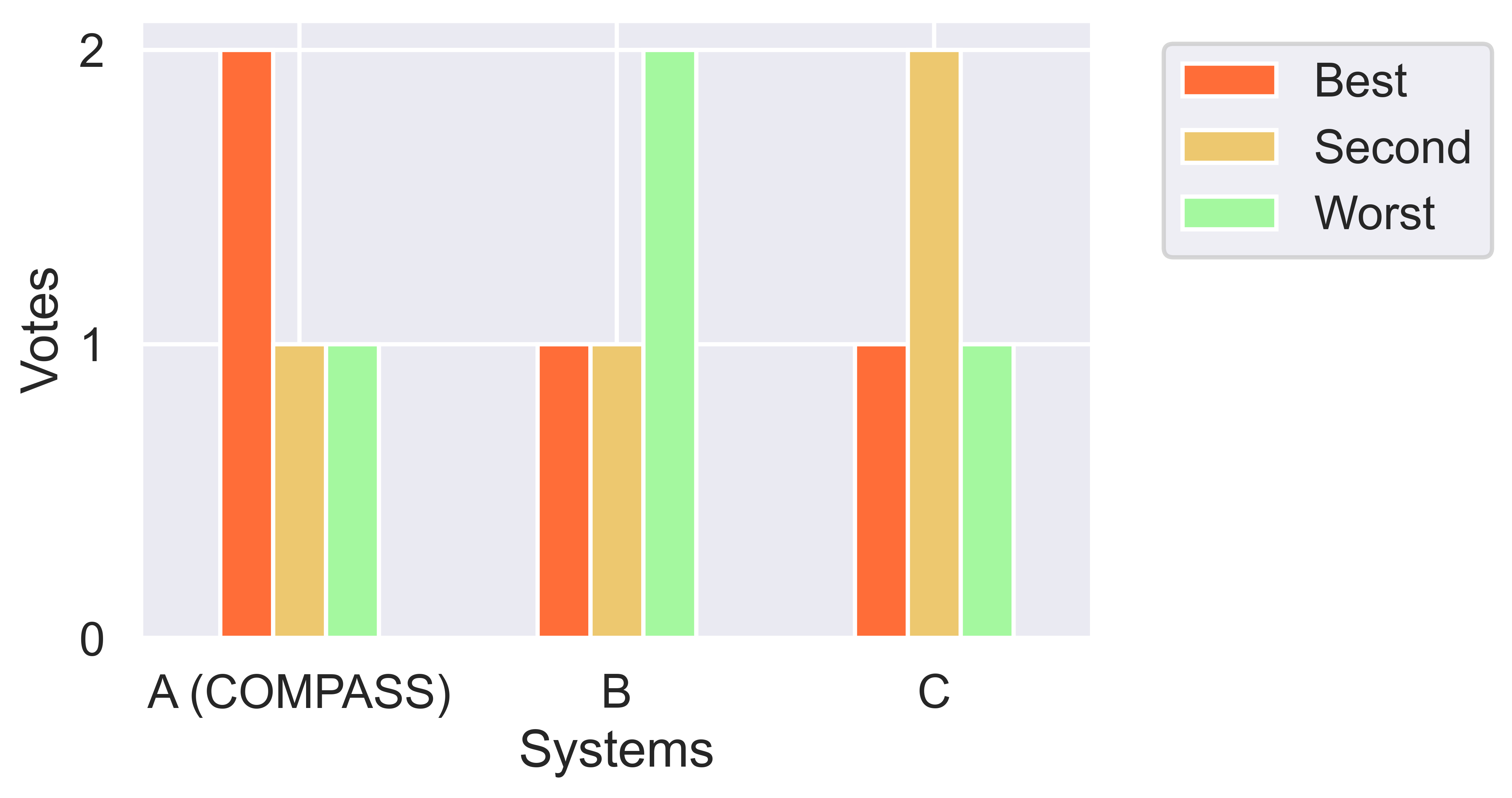}
    \end{center}
    \caption{Bar plot of the results of systems comparison.}
    \label{fig:ranking_bar_plot}
\end{figure}

Table \ref{tab:ranking} shows the collected answers. The responses varied, but overall System A, i.e., our proposed system, received the highest rating. To make the results easier to understand at a glance, a bar plot is presented in Figure \ref{fig:ranking_bar_plot}.

The evaluator who said that C was the best stated that diversity was more important than accuracy for proposing missing positions. Therefore, C, which randomly gave out more missing positions, was probably more suitable than A, which did not point out missing parts unless necessary.
The evaluator who said that B was best emphasized the importance of being able to develop the story, and it is likely that B, which generates a continuation, was more appropriate support for this person than the other systems.

\begin{table}[!t]
\footnotesize
\center
\begin{tabular}{p{2cm}p{2cm}p{2cm}p{9cm}}
\toprule\relax
\textbf{Best} & \textbf{Second} & \textbf{Worst} & \textbf{Reason} \\
\midrule\relax
\textbf{A (COMPASS)} & B & C & システムＡはわりと意味が通る面白い文章が出ました。Ｂ，Ｃとなるうちに、文が捻られいい不明な文脈になっていきました。 (System A produced interesting sentences that made rather good sense, but as System B and C progressed, the sentences became twisted and the context became unclear.) \\
\midrule\relax
\textbf{A (COMPASS)}  & C & B & Bは作品のプラスになる内容が候補文に出てくることが少なく感じた。AとCは甲乙つけがたいがややAの方が面白い提示が多く感じる。 (In the case of B,  I felt that there was not much in the candidate text that was positive for the work. Although it was difficult to decide between A and C, I felt that A had a slightly more interesting presentation.) \\
\midrule\relax
C & \textbf{A (COMPASS)}  & B & 文章によって結果が異なるが、全体的にＣが最も提案の幅があるように感じた。missing sentenceの追加される場所も多く感じた。一方、Ｂは最も淡泊だった。正直、この手のシステムは正確性よりも提案の幅のほうが重要だと感じたため、この順番になった。 (The results differed depending on the text, but overall C seemed to have the widest range of suggestions, with many places where ``missing sentence'' is added. B, on the other hand, was the most bland. To be honest, I felt that the breadth of suggestions was more important than the accuracy in this kind of system, which is why I chose this order.) \\
\midrule\relax
B & C & \textbf{A (COMPASS)} &                             いちばんストーリーに発展性を感じられたから。 (Because I felt it would develop the story the most.) \\
\bottomrule\relax
\end{tabular}
\caption{Ranking of the systems by each evaluator and the reason given for the evaluation.}
\label{tab:ranking}
\end{table}

\subsubsection{Evaluation of Modules}
\label{subsubsec:evaluation_modules}

\begin{figure}[!t]
    \begin{center}
    \includegraphics[clip,width=\columnwidth]{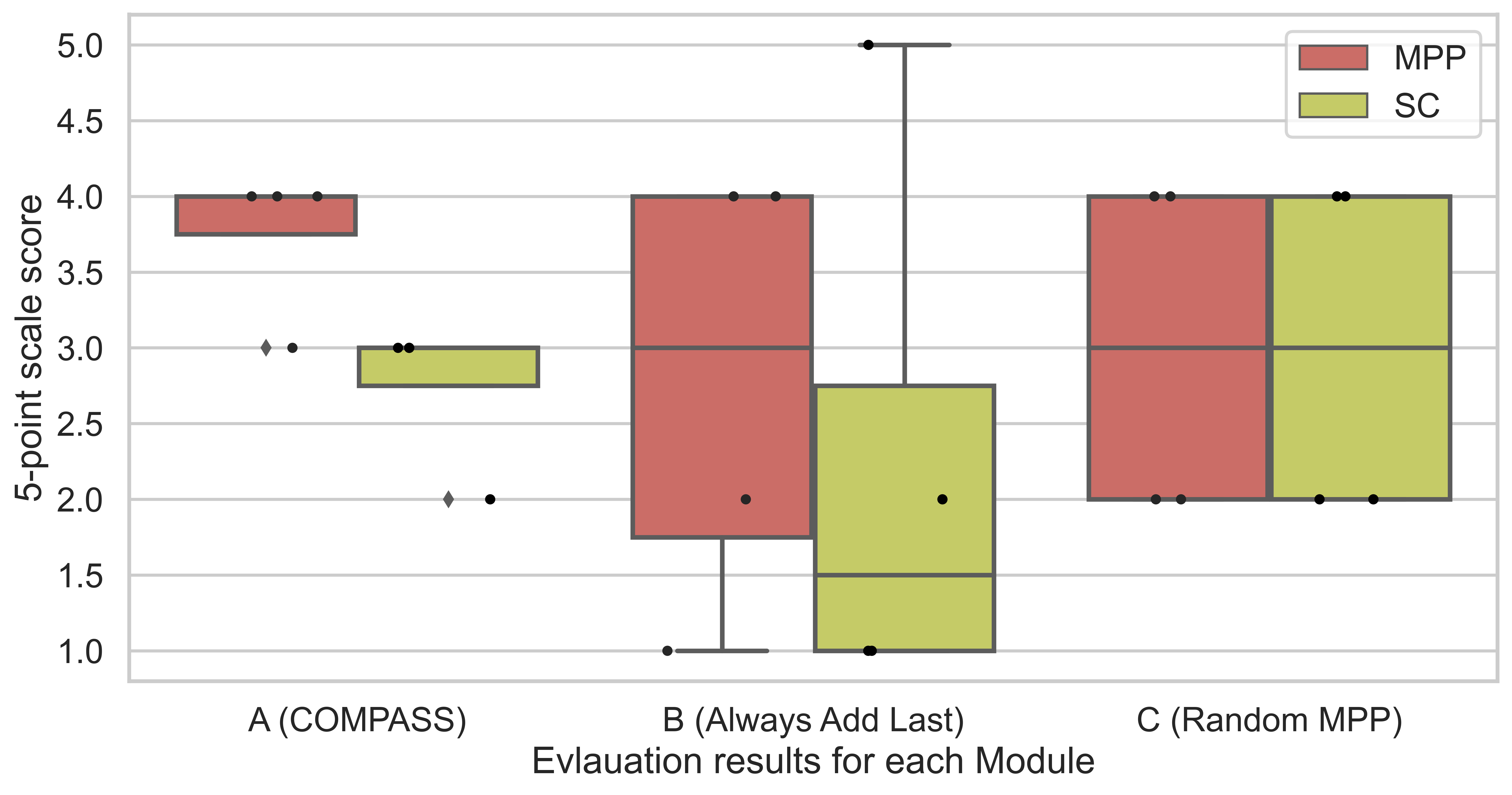}
    \end{center}
    \caption{Box plot of 5-point evaluation results for each module in each system.}
    \label{fig:evaluate_each_module}
\end{figure}

We also asked the evaluators to rate each module of each system on a 5-point scale. The results are shown in Figure \ref{fig:evaluate_each_module}.

Regarding the MPP modules, the proposed method received the highest evaluation. 
It is interesting to note that although the same SC modules were used in the three systems, their evaluations differed significantly.

\subsubsection{Usefulness of the Functions}

\begin{figure}[!t]
    \begin{center}
    \includegraphics[clip,width=\columnwidth]{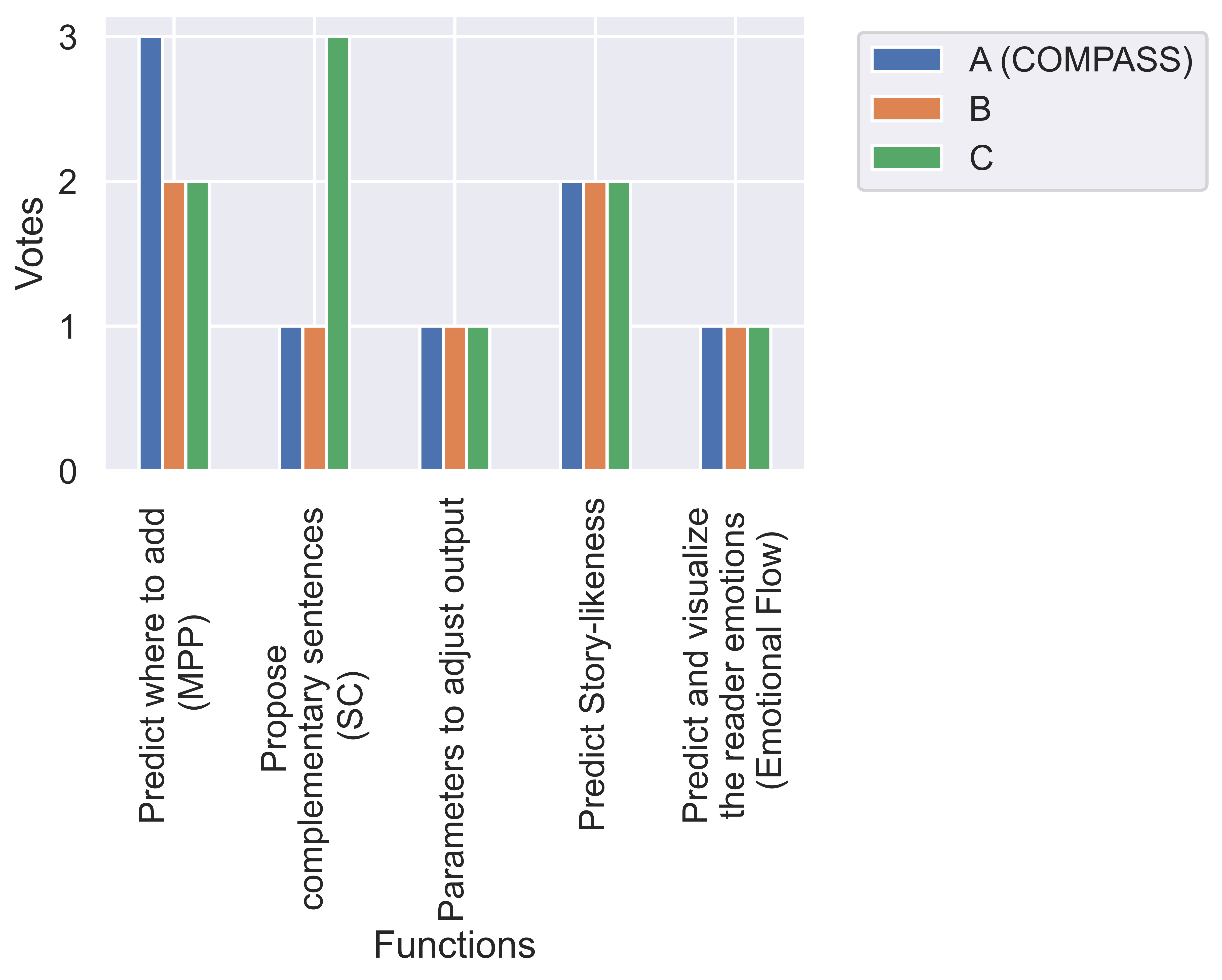}
    \end{center}
    \caption{Bar plot of the responses regarding useful functions.}
    \label{fig:functions_bar_plot}
\end{figure}

For each system, we asked the evaluators about the usefulness of each function. In this question, evaluators were allowed to choose as many functions as they thought to be useful.
They were also asked to write the reason for their answer.
As shown in Table \ref{table:user_study_questionnaire_1}, the choices of functions were as follows: 

\begin{itemize}
    \item Predict where to add -- MPP
    \item Propose complementary sentences -- SC
    \item Parameters to adjust output
    \item Predict Story-likeness
    \item Predict and visualize reader emotions -- Emotional Flow
    \item Others 
\end{itemize}

Although the terms MPP and SC were not communicated to the evaluators, the correspondence with their choices is shown for reference.
For Emotional Flow, the name was given to the evaluator as shown in Table \ref{fig:ja-system_canditate_and_emotional_flow}.

The result of the votes are shown in Figure \ref{fig:functions_bar_plot}, and the reason why they consider functions useful/not useful are shown in Table \ref{tab:functions_comments}.

\begin{table}[!t]
\footnotesize
\center
\begin{tabular}{p{1.6cm}p{12.4cm}}
\toprule
\textbf{System} & \textbf{Reasons to consider functions as useful/not useful} \\
\midrule
A (COMPASS) & 
    \begin{tabular}{p{12.4cm}}
        - ストーリーらしさの数値化は面白かった。加筆すべき箇所はラストに来ることが多かったが、その他の箇所に来ることもあり、自分だったら何を入れるかを検討するのに役だった。補完文の提案内容はつたなかった。 (It was interesting to quantify the story-likeness. The parts that needed to be added were often in the last part of the story, but sometimes they were in other parts, which helped me to think about what I would add. The suggestions for complementary sentences were not so good.) \\ 
        \\
        - 調整用パラメータはいじれば提示も変わってくるものの、どう影響しているのかが分かりにくい。ストーリーらしさの数値化は正直なところそれの持つ意味がよく分からない。何をもってストーリーらしいと評価しているのか等。読者の感情の可視化はもう少し読み解きやすい形式があると助かるのではないか。以上の点は全てのシステムで同じように感じたため、以降の設問においては回答を省略させていただく。 (The parameters for adjustment can be tweaked to change the presentation, but it's hard to see how they affect it. To be honest, I don't really understand the meaning behind the quantification of story-likeness. How the system evaluate story-likeness? It would be helpful if the visualization of the reader's emotions were in an easier format to read and understand. Since I felt the same way about the above points in all systems, I will omit my answers to the following questions.)\\ 
        \\
        - それが一文だけでなく長期的に視覚化できるようになると、物語構築が立てやすい。  (If it can be visualized in the long term, not just in one sentence, it is easier to build the narrative structure.) (* ``it (それ)'' in this comment refers to ``Predict and visualize the reader emotions.'') \\
    \end{tabular} \\
\midrule
B &     
    \begin{tabular}{p{12.4cm}}
        - 感情変化の可視化は作家にとって有用な情報。 (Visualization of emotional changes is useful information for writers.)\\
        \\
        - ストーリーらしさの数値化は、評価のクオリティが高まれば第三者的な視点として役に立つと思う。 (I think the quantification of story-likeness can be useful as a third party perspective if the quality of the evaluation is high enough.)\\
    \end{tabular} \\ 
\midrule
C &    
    \begin{tabular}{p{12.4cm}}
        - 感情は有益な情報。 (Emotions are useful information.)\\
        \\
        - 省略すると言いつつ最後に一つ思い出したので書いておくと、自分はマウスのホイールでページを上下に送ろうとするタイプで、マウスカーソルが読者感情の可視化のグラフのところにある時にその動作をしようとするとグラフが拡大縮小されてしまいページを上下出来ないことに若干の不便を感じた。 (I'm the type of person who tries to move the page up and down with the mouse wheel, and when the mouse cursor is on the graph of reader emotions visualization, the graph is scaled up and down and I can't move the page up and down. I found this slightly inconvenient.)\\
        \\
        - 加筆すべき箇所の提案が多かったのは、自分が改めて「そこになにか入れられないか？」と考え直すきっかけに出来るように思う。補完文に関してはそのまま活かせる印象はなかった。 (There were a lot of suggestions of the place to add some text, which I think will make me rethink, "What can I put in there?" I didn't have the impression that I could make use of the complementary sentences as it is.) \\
    \end{tabular} \\
\bottomrule
\end{tabular}
\caption{Reasons given by evaluators for selecting the functions as useful for story creation.}
\label{tab:functions_comments}
\end{table}

From the reasons given the evaluators, it is clear that different evaluators had different thoughts about the functions.
Some of evaluators found the quantification of story-likeness interesting, whereas others were uncomfortable with the fact that it was not clear how it was being evaluated.
In addition, some of them focused on emotion visualization; we anticipate that our Emotional Flow will be able to meet the needs of such users when valence and arousal estimation become more accurate.

On the other hand, the complementary sentences were generally rated low. This suggests that, from a professional novelist's point of view, the model trained on the current dataset does not generate sufficiently good sentences.

\subsection{Future Direction}

\subsubsection{Copyright and Privacy}
\label{subsubsec:copyright_and_privacy}

In the evaluation experiment, the proposed system and the comparison systems were executed on the (virtual) server managed by us, who conducted the experiment. 
This is because we avoided sharing the source code with the evaluators in order to prevent them from knowing which of the three systems, A, B, and C, was the proposed system.
However, when considering the operation of the proposed system in the real world, 
running it as a stand-alone system in the hands of each user is an important topic to consider.

Privacy is an important issue in the exchange of information via the Internet.
In particular, in the case of creative writing support systems, the nature of supporting the writing of works that have not yet been released to the world means that copyright must be carefully handled alongside with privacy.
When the system operator collects the creative works input by the users, it is necessary to establish rules to protect the users' rights, and the users must be fully convinced and assured that their rights are being protected.
Alternatively, one possible solution is to make the system a stand-alone system where no one but the user oneself can see the input.
Our proposed system has also been verified to work on CPUs. It can be run in environments that are not rich in computing resources, although the response time experienced by the user will be longer.

\subsubsection{Expansion of Japanese language data}
\label{subsubsec:japanese_data}

Although it may conflict with the considerations mentioned above of privacy and copyright, gathering Japanese story data and constructing a significant and high-quality dataset is an essential part of the system's future development.

In this study, we developed a Japanese version of the system in order to have professional creators who work in Japanese evaluate the system. 
However, there was a major difficulty in doing so, especially regarding the lack of datasets.

As an alternative approach, we have successfully trained a module of MPP by translating the entire English novel dataset into Japanese.
However, it was also confirmed that applying the same method to the SC module was undesirable because it would destroy the style and nuance of the text.

In fact, some of the evaluators complained about the generated complementary sentences as shown in Table \ref{tab:functions_comments}.
We should consider improving the method to learn better sentence expressions from a small amount of data; however, it is also important to develop high-quality and large-scale data sets.

The understanding and cooperation of the people who actually produce such data, i.e., professional creators, is essential for the realization of such high-quality and large-scale data sets.

\subsubsection{Personalize}
\label{subsubsec:personalzize}

In the evaluation experiment, all evaluators were asked to try the same three systems.
As a result, it became clear that each evaluator preferred a different system.
Overall, the proposal method was the best, but the evaluation was uneven.
What each evaluator wants from the creation support system is also in a different direction.
What is desirable for one author may not necessarily be desirable for another author.
Therefore, \textbf{personalization} will play an important role in the future direction of creation support systems.

We did not conduct a long-term experiment because it would have been a burden on the subjects.
However, it is possible to adapt the output of the system to the user based on the user's input data and feedback on the output.
Further fine-tuning and automatic optimization of the parameters for output adjustment may be considered.

With the user's permission, if the system can use the person's previous writings as training data, it is expected to be able to generate sentences that are more like the user's own writing style.

\section{Conclusion}

To overcome the issue of conventional SC tasks requiring information regarding the position of the missing part in a story, we previously proposed an MPP that predicts the position based on the given incomplete story.
Specifically, we proposed MPP with limited conditions (LC-MPP) in \citep{mori-etal-2020-finding}. 
In this paper we proposed an updated version called Variable Number MPP (VN-MPP).

In LC-MPP, it is known that there is a missing position in the input story, and that there is only one such instance.
However, in reality, an input story may be complete, that is, $k$ is null. Furthermore, there may be a case in which there are multiple missing positions, that is, a case in which $k$ has multiple values.
We thus proposed VN-MPP as a task closer to a more realistic setting to address this issue.
Furthermore, we proposed two novel methods for VN-MPP and Story Completion (SC): the two-module approach and the end-to-end approach.
Our proposed method not only copes with the variability in the number of missing sentences, but also with the variability in the number of sentences in the input.

Based on our proposed MPP task, we developed a story writing support system, \textbf{COMPASS}.
Further, we created a Japanese version of the system and conducted an evaluation experiment involving professional evaluators who are currently engaged in creative activities.
The results obtained confirm the usefulness of our proposed MPP-based writing support system.

By having the evaluators actually use the system, we were able to obtain concrete answers regarding their desiderata for creative support systems. Based on their feedback, we will further develop the system to make it more useful for creative support.

\section{Acknowledgements}
\label{acknowledgements}

This work was partially supported by JST AIP Acceleration Research JPMJCR20U3, Moonshot R\&D Grant Number JPMJPS2011, JSPS KAKENHI Grant Number JP20H05556 and Basic Research Grant (Super AI) of Institute for AI and Beyond of the University of Tokyo.

\bibliographystyle{abbrvnat}

\end{document}